\documentclass[11pt]{article}

\usepackage[preprint]{acl}
\usepackage{pifont}
\usepackage{times}
\usepackage{latexsym}

\usepackage[T1]{fontenc}

\usepackage[utf8]{inputenc}

\usepackage{microtype}

\usepackage{inconsolata}

\usepackage{graphicx}

\usepackage{multirow}
\usepackage[most]{tcolorbox}

\tcbset{
  promptbox/.style={
    width=\columnwidth,
    colback=gray!5,
    colframe=gray!40,
    coltitle=black,
    fonttitle=\bfseries,
    boxrule=0.4pt,
    arc=4pt,
    left=4pt,
    right=4pt,
    top=4pt,
    bottom=4pt,
    fontupper=\normalsize
  }
}

\newcommand{\sym}[1]{\ensuremath{^{#1}}}

\usepackage[most]{tcolorbox}
\usepackage{booktabs}
\usepackage{siunitx}
\usepackage{makecell}
\usepackage[table]{xcolor}
\usepackage{tcolorbox}
\usepackage{tabularx}
\usepackage{soul}
\newcommand{\hlg}[1]{{\sethlcolor{green!20}\hl{#1}}}
\newcommand{\hlb}[1]{{\sethlcolor{blue!20}\hl{#1}}}

\title{How Robust Is Multimodal Claim Verification to LLM Rewriting?}

\author{
Yun-Ang Wu,$^1$ 
Xanh Ho,$^1$ 
Andre Greiner-Petter,$^{2,3}$ 
Sunisth Kumar,$^4$ \\
\textbf{Tian Cheng Xia},$^5$ 
\textbf{Florian Boudin},$^6$ 
\textbf{Akiko Aizawa}$^{1,2,4}$ \\ 
$^1$NII LLMC, Japan \hspace{1cm}
$^2$National Institute of Informatics, Japan \hspace{1cm}\\
$^3$University of Göttingen, Germany 
$^4$The University of Tokyo, Japan \hspace{1cm}\\
$^5$University of Bologna, Italy \hspace{1cm}
$^6$Inria, LS2N, Nantes Université, France \\
{\tt \{yunangwu, xanh, aizawa\}@nii.ac.jp} \hspace{1cm}
{\tt greinerpetter@gipplab.org} \\
{\tt sunisth@g.ecc.u-tokyo.ac.jp} \hspace{1cm}
{\tt tiancheng.xia@studio.unibo.it} \\
{\tt florian.boudin@univ-nantes.fr}
}

\begin{document}
\maketitle

\begin{abstract}
LLMs are known to introduce stylistic changes into generated text, yet how these stylistic shifts affect model decisions on scientific tasks remains underexplored. In this paper, we focus on multimodal claim verification, where the goal is to determine whether a textual claim is grounded in a given piece of evidence. We apply two rewriting strategies: natural rewriting, which simulates how researchers routinely use LLMs to polish academic text, and controlled injection, which inserts a single LLM-associated word to isolate the effect of vocabulary choice. We evaluate 11 open-weight models spanning five VLM families and ranging from 2B to 38B parameters. We find that models are robust to these modifications: most show no significant drop in accuracy, and compared to prior work on review-score manipulation, verification appears far more stable. However, consistent probability shifts do occur. Hedging-oriented conditions produce significant shifts across nearly all models, while boosting conditions show a weaker effect and general polishing conditions (e.g., grammar correction, fluency improvement) have little effect.\footnote{\url{https://github.com/yunangwu/llm-rewrite-claim-verification}}

\end{abstract}

\section{Introduction}

\begin{figure}[t]
\centering
\includegraphics[width=\columnwidth]{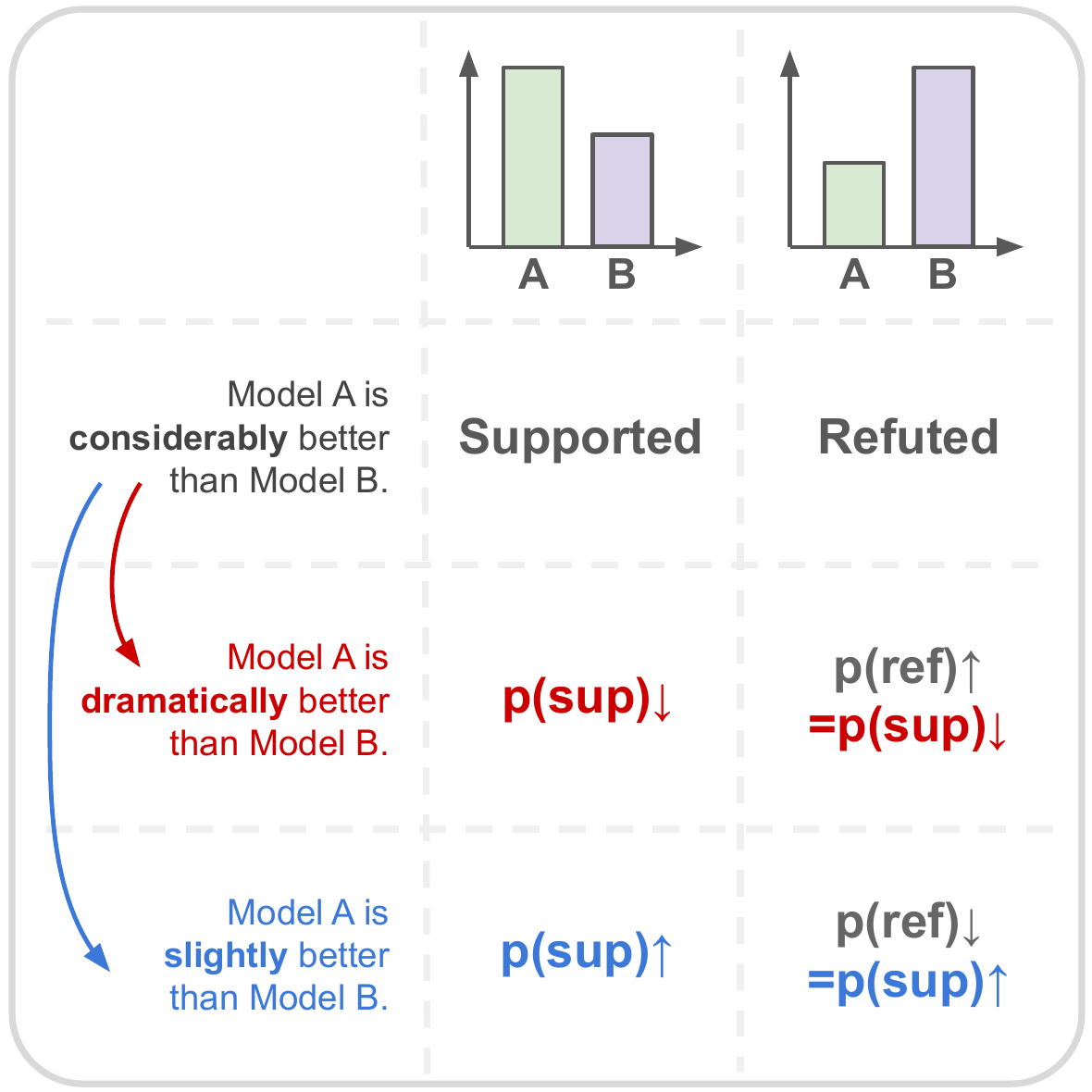}
\caption{A claim verification task requires deciding whether a claim is supported or refuted by evidence. This figure shows how a rational decision-maker should adjust $p(\text{support})$ when the claim's intensity modifier changes. A stronger modifier (e.g., ``considerably'' to ``dramatically'') makes a supported claim harder to verify and a refuted claim easier to refute, both lowering $p(\text{support})$. However, this reasoning only applies when the modification is semantically meaningful. Our central question: when the modification is stylistic and does not alter the claim's semantic content, does the model still exhibit the same behavioral shift?}
\label{fig:claim_modification}
\end{figure}

LLMs are quietly shaping how scientific papers are written and reviewed. Several studies \citep{doi:10.1126/sciadv.adt3813, Matsui2025-nj, cunningham2025shiftingnormsscholarlypublications} report that the quality and style of scientific writing have changed since the appearance of ChatGPT.
At the same time, conferences such as ICLR have started exploring LLM-assisted review pipelines \citep{thakkar2025llmfeedbackenhancereview}, suggesting that LLMs may increasingly be involved in evaluating text that LLMs helped produce.

These stylistic shifts may have consequences beyond presentation, affecting how downstream systems evaluate scientific text. Indeed, recent work shows that adversarial paraphrasing and textual adversarial attacks can inflate paper-level review scores \citep{kaneko2026paraphrasingadversarialattackllmasareviewer,lin-etal-2025-breaking}. But paper-level reviewing conflates many factors (e.g., argument structure, novelty, presentation), making it hard to isolate what drives the effect. We instead study single-claim verification against visual evidence, where the only variable is how the claim is worded.

We study this through multimodal claim verification~\citep{ho2026sciclaimevalcrossmodalclaimverification, wang-etal-2025-sciver, lal2025musciclaimsmultimodalscientificclaim}, where a textual claim is checked against a table or figure from a scientific paper. Using SciClaimEval \citep{ho2026sciclaimevalcrossmodalclaimverification}, we rewrite claims under two strategies: \textit{natural rewriting}, which paraphrases claims using LLM prompts, and \textit{controlled injection}, which inserts individual words into otherwise unchanged claims. Natural rewriting simulates five scenarios that reflect how researchers routinely use LLMs (e.g., grammar correction, fluency improvement). Controlled injection isolates the effect of single words, showing that even one vocabulary insertion can shift a model's prediction.

We measure how model confidence changes by modeling $p(\text{support})$ as the average of 10 runs using self-consistency~\citep{wang2023selfconsistencyimproveschainthought} and test on 11 models across 5 model families. We find that models are robust to stylistic change in terms of accuracy, regardless of family or size. On the other hand, we observe meaningful probability shifts in a consistent direction under different rewriting strategies.

\begin{table*}[t]
\centering
\small
\begin{tabularx}{\linewidth}{lX}
\toprule
\textbf{Condition} & \textbf{Claim} \\
\midrule
Original & Additionally, 46.6\% of participants were primary school graduates, 9.4\% reported alcohol use, and 13\% were smokers, with an average duration of smoking of 22.83 ± 6.94 years. \\
\midrule
\multicolumn{2}{l}{\textit{Natural Rewriting}} \\
\midrule
Language Polishing (LP) & Additionally, \hlg{46.6 \%} of participants were \hlg{primary‑school} graduates, \hlg{9.4 \%} reported alcohol use, and \hlg{13 \%} were smokers, with an average \hlg{smoking} duration of 22.83 ± 6.94 years. \\
Fluency Improvement (FI) & \hlg{Moreover, 46.6 \%} of \hlg{the} participants \hlg{had completed} primary \hlg{school; 9.4 \%} reported \hlg{consuming alcohol;} and \hlg{13 \%} were smokers, with an average smoking \hlg{history} of 22.83 ± 6.94 years. \\
Academic Style Alignment (AS) & \hlg{Furthermore, 46.6 \%} of participants \hlg{had completed} primary \hlg{education, 9.4 \%} reported alcohol \hlg{consumption,} and \hlg{13 \%} were smokers, with \hlg{a mean smoking} duration of 22.83 ± 6.94 years. \\
Outcome-Aware (OA) & \hlg{Notably, 46.6 \%} of \hlg{the cohort} were \hlg{primary‑school} graduates, \hlg{9.4 \%} reported alcohol \hlg{consumption,} and \hlg{13 \%} were smokers, with \hlg{a mean} smoking \hlg{history} of 22.83 ± 6.94 years. \\
Cautious Tone (CT) & Additionally, \hlg{approximately} 46.6\% of participants \hlg{had completed} primary \hlg{school, about} 9.4\% reported alcohol \hlg{consumption,} and \hlg{roughly} 13\% \hlg{identified as} smokers, with \hlg{a reported} average \hlg{smoking} duration of 22.83 ± 6.94 years. \\
\midrule
\multicolumn{2}{l}{\textit{Controlled Injection}} \\
\midrule
Conservative (C1) & Additionally, \hlb{approximately} 46.6\% of participants were primary school graduates, 9.4\% reported alcohol use, and 13\% were smokers, with an average duration of smoking of 22.83 ± 6.94 years. \\
Decorative (D1) & Additionally, \hlb{noteworthy} 46.6\% of participants were primary school graduates, 9.4\% reported alcohol use, and 13\% were smokers, with an average duration of smoking of 22.83 ± 6.94 years. \\
\bottomrule
\end{tabularx}
\caption{Rewrite examples for claim \texttt{val\_tab\_0484} from SciClaimEval \citep{ho2026sciclaimevalcrossmodalclaimverification}. Natural rewriting paraphrases claims using LLM prompts that simulate common academic editing scenarios, while controlled injection inserts a single LLM-associated word into the original claim.}
\label{tab:rewrite-example}
\end{table*}

The core contributions of this paper are listed as follows:
\begin{itemize}
    \item A systematic empirical study of stylistic sensitivity in multimodal claim verification. We evaluate 11 open-weight VLMs across five families under two rewriting strategies (natural LLM paraphrasing and controlled single-word injection), controlling for stylistic variation as the primary experimental variable.
    \item We show that several rewriting conditions, especially cautious-tone and conservative-word injection, induce consistent probability shifts even when accuracy remains stable, revealing a more nuanced vulnerability than what has been previously documented for other scientific tasks such as paper reviewing~\citep{kaneko2026paraphrasingadversarialattackllmasareviewer,lin-etal-2025-breaking}.
\end{itemize}

\section{Related Work}
\subsection{LLM-Assisted Scientific Writing}
Several studies have examined the use of LLMs as writing tools and the traits of LLM-assisted text. \citet{pmlr-v235-liang24b} proposed a method for estimating the fraction of LLM-modified content in a corpus and applied it to peer reviews from AI conferences held after the release of ChatGPT, estimating, for example, that 16.9\% of review sentences at EMNLP 2023 were substantially modified by LLMs. \citet{liang2024mapping} applied this framework to 950k papers from arXiv, bioRxiv, and Nature portfolio journals published between 2020 and 2024, observing a steady increase in LLM usage, with computer science showing the largest growth.

\citet{doi:10.1126/sciadv.adt3813} examined 15 million PubMed abstracts and concluded that at least 13.5 percent of those published in 2024 were generated or written with the assistance of an LLM, with certain style words such as ``crucial,'' ``delve,'' and ``potential'' being used excessively. \citet{Matsui2025-nj} compiled a list of LLM style words from previous research and compared them against a control group, finding that 103 out of 135 words showed meaningful uplift in the post-LLM era. \citet{cunningham2025shiftingnormsscholarlypublications} analyzed 17 million papers from Semantic Scholar and reported that in fields such as computer science and engineering, writing styles have trended toward ``hyperbole'' since 2023.

\subsection{LLM-Assisted Adversarial Rewriting}
There is also work showing that rewriting or injecting words, even without changing its factual content, can influence the model's review score. \citet{lin-etal-2025-breaking} proposed that textual adversarial attacks can artificially inflate LLM review scores while keeping the semantic content of the paper unchanged. Different attack methods (character-level, word-level, and sentence-level) all demonstrate some degree of effectiveness. \citet{kaneko2026paraphrasingadversarialattackllmasareviewer} instead employed an LLM-based paraphrasing approach as the main attack method and reported that this method can optimize the review score without changing the claims of the paper.

\subsection{Multimodal Claim Verification}
\begin{table*}[t]
\centering
\footnotesize
\setlength{\tabcolsep}{4pt}
\resizebox{\textwidth}{!}{%
\begin{tabular}{l l c ccccccc}
\toprule
 & & & \multicolumn{5}{c}{\textbf{Natural Rewriting}} & \multicolumn{2}{c}{\textbf{Controlled Rewriting}} \\
\cmidrule(lr){4-8} \cmidrule(lr){9-10}
\textbf{Model} & & \textbf{Base} & LP & FI & AS & OA & CT & D1 & C1 \\
\midrule
 \multirow{2}{*}{gemma-4-26B-A4B-it} & P-Acc & 0.548 & \cellcolor{red!5}0.546 & \cellcolor{red!9}0.537 & \cellcolor{red!15}0.529 & \cellcolor{red!13}0.532 & \cellcolor{red!19}0.524 & \cellcolor{green!9}0.560 & \cellcolor{red!7}0.538 \\
  & F1 & 0.789 & \cellcolor{red!5}0.785 & \cellcolor{red!5}0.782 & \cellcolor{red!13}0.773 & \cellcolor{red!15}0.770 & \cellcolor{green!5}0.790 & \cellcolor{green!5}0.792 & \cellcolor{green!5}0.790 \\
\addlinespace[2pt]
 \multirow{2}{*}{gemma-4-31B-it} & P-Acc & 0.610 & \cellcolor{red!20}0.585 & \cellcolor{red!25}0.579\textsuperscript{*} & \cellcolor{red!23}0.581 & \cellcolor{red!36}0.565\textsuperscript{***} & \cellcolor{red!40}0.557\textsuperscript{***} & \cellcolor{red!5}0.606 & \cellcolor{red!19}0.586 \\
  & F1 & 0.824 & \cellcolor{red!7}0.814 & \cellcolor{red!11}0.809\textsuperscript{*} & \cellcolor{red!11}0.810 & \cellcolor{red!20}0.798\textsuperscript{***} & \cellcolor{red!14}0.806\textsuperscript{*} & \cellcolor{red!5}0.822 & \cellcolor{red!6}0.816 \\
\addlinespace[2pt]
 \multirow{2}{*}{GLM-4.6V-Flash} & P-Acc & 0.571 & \cellcolor{red!5}0.566 & \cellcolor{red!6}0.563 & \cellcolor{red!5}0.565 & \cellcolor{red!6}0.563 & \cellcolor{red!25}0.539 & \cellcolor{red!5}0.565 & \cellcolor{red!5}0.568 \\
  & F1 & 0.791 & \cellcolor{red!5}0.791 & \cellcolor{red!5}0.788 & \cellcolor{red!5}0.790 & \cellcolor{red!5}0.785 & \cellcolor{red!5}0.790 & \cellcolor{green!5}0.792 & \cellcolor{green!5}0.797 \\
\addlinespace[2pt]
 \multirow{2}{*}{InternVL3-2B} & P-Acc & 0.041 & \cellcolor{green!13}0.057 & \cellcolor{green!9}0.052 & \cellcolor{green!9}0.052 & \cellcolor{green!15}0.060 & \cellcolor{red!5}0.037 & \cellcolor{green!8}0.051 & \cellcolor{green!5}0.045 \\
  & F1 & 0.658 & \cellcolor{green!5}0.664 & \cellcolor{green!5}0.664 & \cellcolor{green!5}0.665 & \cellcolor{green!5}0.665 & \cellcolor{green!7}0.667 & \cellcolor{green!6}0.666 & \cellcolor{green!6}0.667 \\
\addlinespace[2pt]
 \multirow{2}{*}{InternVL3-8B} & P-Acc & 0.208 & \cellcolor{red!14}0.190 & \cellcolor{red!14}0.190 & \cellcolor{red!17}0.187 & \cellcolor{red!12}0.193 & \cellcolor{red!40}0.135\textsuperscript{***} & \cellcolor{red!5}0.203 & \cellcolor{red!28}0.173 \\
  & F1 & 0.699 & \cellcolor{red!6}0.691 & \cellcolor{red!5}0.696 & \cellcolor{red!5}0.692 & \cellcolor{red!5}0.692 & \cellcolor{red!9}0.687 & \cellcolor{green!5}0.700 & \cellcolor{red!5}0.695 \\
\addlinespace[2pt]
 \multirow{2}{*}{InternVL3-38B} & P-Acc & 0.338 & \cellcolor{red!33}0.296 & \cellcolor{red!24}0.307 & \cellcolor{red!38}0.289 & \cellcolor{red!35}0.294 & \cellcolor{red!40}0.273 & \cellcolor{green!5}0.339 & \cellcolor{red!7}0.329 \\
  & F1 & 0.718 & \cellcolor{red!11}0.704 & \cellcolor{red!14}0.701 & \cellcolor{red!15}0.699 & \cellcolor{red!20}0.692 & \cellcolor{red!7}0.709 & \cellcolor{red!5}0.712 & \cellcolor{green!5}0.718 \\
\addlinespace[2pt]
 \multirow{2}{*}{Kimi-VL-A3B} & P-Acc & 0.353 & \cellcolor{green!7}0.362 & \cellcolor{green!38}0.401\textsuperscript{***} & \cellcolor{green!20}0.378 & 0.353 & \cellcolor{red!10}0.340 & \cellcolor{green!5}0.358 & \cellcolor{red!10}0.340 \\
  & F1 & 0.730 & \cellcolor{green!5}0.733 & \cellcolor{green!7}0.740 & \cellcolor{green!6}0.739 & \cellcolor{red!5}0.725 & \cellcolor{red!5}0.727 & \cellcolor{red!5}0.725 & \cellcolor{red!5}0.726 \\
\addlinespace[2pt]
 \multirow{2}{*}{Qwen3-VL-2B-Instruct} & P-Acc & 0.259 & \cellcolor{green!21}0.286 & \cellcolor{green!11}0.273 & \cellcolor{green!5}0.260 & \cellcolor{green!12}0.274 & \cellcolor{red!5}0.254 & \cellcolor{green!5}0.264 & \cellcolor{red!5}0.257 \\
  & F1 & 0.658 & \cellcolor{green!6}0.667 & \cellcolor{green!5}0.663 & \cellcolor{red!5}0.658 & \cellcolor{green!8}0.669 & \cellcolor{green!5}0.663 & \cellcolor{green!5}0.661 & \cellcolor{green!5}0.664 \\
\addlinespace[2pt]
 \multirow{2}{*}{Qwen3-VL-4B-Instruct} & P-Acc & 0.452 & \cellcolor{green!5}0.453 & \cellcolor{green!8}0.462 & \cellcolor{red!5}0.447 & \cellcolor{green!17}0.473 & \cellcolor{green!28}0.487 & \cellcolor{green!12}0.468 & \cellcolor{green!9}0.464 \\
  & F1 & 0.710 & \cellcolor{green!5}0.717 & \cellcolor{green!7}0.719 & \cellcolor{red!5}0.707 & \cellcolor{green!5}0.716 & \cellcolor{green!27}0.745\textsuperscript{***} & \cellcolor{green!11}0.725 & \cellcolor{green!11}0.725 \\
\addlinespace[2pt]
 \multirow{2}{*}{Qwen3-VL-8B-Instruct} & P-Acc & 0.516 & \cellcolor{green!10}0.529 & \cellcolor{green!5}0.518 & \cellcolor{red!5}0.513 & \cellcolor{red!5}0.513 & \cellcolor{green!21}0.543 & \cellcolor{green!5}0.519 & \cellcolor{red!12}0.501 \\
  & F1 & 0.739 & \cellcolor{green!7}0.748 & \cellcolor{green!5}0.740 & \cellcolor{green!5}0.740 & \cellcolor{red!10}0.725 & \cellcolor{green!21}0.765 & \cellcolor{green!5}0.741 & \cellcolor{red!5}0.738 \\
\addlinespace[2pt]
 \multirow{2}{*}{Qwen3-VL-32B-Instruct} & P-Acc & 0.617 & \cellcolor{red!6}0.609 & \cellcolor{green!5}0.618 & \cellcolor{red!22}0.589 & \cellcolor{red!31}0.577 & \cellcolor{red!10}0.604 & \cellcolor{red!5}0.610 & \cellcolor{green!5}0.619 \\
  & F1 & 0.805 & \cellcolor{red!5}0.800 & \cellcolor{red!5}0.803 & \cellcolor{red!13}0.788 & \cellcolor{red!27}0.771\textsuperscript{***} & \cellcolor{green!5}0.808 & \cellcolor{red!5}0.798 & \cellcolor{green!5}0.810 \\
\bottomrule
\end{tabular}
}
\caption{Claim verification performance under different rewriting conditions. \textbf{Base}: original unmodified claims. Significance levels: * $q < 0.05$, ** $q < 0.01$, *** $q < 0.001$ (BH-corrected).\protect\footnotemark}
\label{tab:main_results}
\end{table*}

Claim verification is the task of determining whether a claim is supported by given evidence (classification). Datasets such as FEVER and SciFact \citep{thorne-etal-2018-fever, wadden-etal-2020-fact} are well-known benchmarks for this task. In this paper, we focus on a specific subset: multimodal scientific claim verification, where the evidence is a figure or table from a paper and the model must ground a claim using the image. SciVer \citep{wang-etal-2025-sciver} collects figures from arXiv computer science papers, and experts were recruited to create both supported and refuted claims based on the evidence. MuSciClaim \citep{lal2025musciclaimsmultimodalscientificclaim} gathers evidence and supported claims from several open-access sources across multiple domains, including physics and biology, and generates refuted claims through manual perturbation. SciClaimEval \citep{ho2026sciclaimevalcrossmodalclaimverification} is another multimodal claim verification dataset that collects diverse evidence from various sources; however, it creates refuted pairs by modifying the evidence rather than the claim.

\subsection{Sensitivity to Surface-Level Perturbations}
LLMs are well known to be affected by surface-level linguistic traits, which can influence their performance on downstream tasks. \citet{xu2023llm} report that LLMs can rewrite a sentence in a meaning-preserving way that changes the predicted label in a classification task, such as sentiment analysis. \citet{sclar2024quantifyinglanguagemodelssensitivity} also claim that LLMs are sensitive to prompt format, with different models reacting differently to variations in prompts across tasks. Furthermore, \citet{chen-etal-2024-humans} show that LLM outputs can be affected by semantic-agnostic biases, such as the inclusion of fake references or superficial stylistic changes. Our work differs from the existing literature in that we focus on how non-adversarial, naturalistic rewriting affects LLMs’ claim verification ability.

\section{Methodology}

\subsection{Multimodal Claim Verification}

Claim verification is the task of determining whether a claim is supported by a given piece of evidence. In the multimodal setting, the evidence is a table or figure image from a scientific paper. Given a claim $c$ and evidence $e$, the goal is to predict the verdict $V(c, e) \in \{\text{support}, \text{refute}\}$. In this work, we perturb the claim by rewriting it using an LLM with a specific rewriting procedure, resulting in a rewritten claim $c'$. We then compare the verdicts $V(c, e)$ and $V(c', e)$ to measure how rewriting affects verification outcomes.

\subsection{Rewriting Strategies}
\label{sec:rewriting}

To investigate how LLM-assisted rewriting affects downstream claim verification, we design two complementary rewriting strategies. The first is \textit{natural rewriting}, which rewrites a claim using a collection of prompts designed to simulate realistic rewriting use cases. The second is \textit{controlled rewriting}, which injects a single word into the claim to study the effect of a small, controlled rewriting operation. An example of how a claim is modified using these rewriting strategies is shown in Table~\ref{tab:rewrite-example}.
\footnotetext{Due to computational constraints, dev set results for InternVL3-38B were not available.}
\subsubsection{Natural Rewriting}
\label{sec:natural_rewriting}
\begin{table*}[t]
\centering
\footnotesize
\setlength{\tabcolsep}{4pt}
\resizebox{\textwidth}{!}{%
\begin{tabular}{l l c l ccccccc}
\toprule
 & & & & \multicolumn{5}{c}{\textbf{Natural Rewriting}} & \multicolumn{2}{c}{\textbf{Controlled Rewriting}} \\
\cmidrule(lr){5-9} \cmidrule(lr){10-11}
\textbf{Model} & & \textbf{Base} & & LP & FI & AS & OA & CT & D1 & C1 \\
\midrule
 \multirow{2}{*}{gemma-4-26B-A4B-it} & $p_{\text{sup}}$ & 0.851 & $\Delta p_{\text{sup}}$ & \cellcolor{red!5}-0.011 & \cellcolor{red!9}-0.024\textsuperscript{***} & \cellcolor{red!9}-0.024\textsuperscript{***} & \cellcolor{red!18}-0.045\textsuperscript{***} & \cellcolor{green!14}+0.037\textsuperscript{***} & \cellcolor{red!5}-0.011\textsuperscript{*} & \cellcolor{green!6}+0.016\textsuperscript{*} \\
  & $p_{\text{ref}}$ & 0.321 & $\Delta p_{\text{ref}}$ & \cellcolor{green!5}+0.007 & \cellcolor{red!5}-0.004 & \cellcolor{green!5}+0.002 & \cellcolor{red!7}-0.019\textsuperscript{*} & \cellcolor{green!27}+0.070\textsuperscript{***} & \cellcolor{red!5}-0.013\textsuperscript{*} & \cellcolor{green!8}+0.021\textsuperscript{***} \\
\addlinespace[2pt]
 \multirow{2}{*}{gemma-4-31B-it} & $p_{\text{sup}}$ & 0.919 & $\Delta p_{\text{sup}}$ & \cellcolor{red!5}-0.002 & \cellcolor{red!6}-0.016\textsuperscript{***} & \cellcolor{red!8}-0.020\textsuperscript{**} & \cellcolor{red!16}-0.040\textsuperscript{***} & \cellcolor{green!7}+0.019\textsuperscript{**} & \cellcolor{red!5}-0.007 & \cellcolor{green!5}+0.005 \\
  & $p_{\text{ref}}$ & 0.315 & $\Delta p_{\text{ref}}$ & \cellcolor{green!9}+0.024\textsuperscript{***} & \cellcolor{green!8}+0.022\textsuperscript{***} & \cellcolor{green!6}+0.015\textsuperscript{*} & \cellcolor{green!5}+0.006 & \cellcolor{green!31}+0.078\textsuperscript{***} & \cellcolor{green!5}+0.010 & \cellcolor{green!14}+0.036\textsuperscript{***} \\
\addlinespace[2pt]
 \multirow{2}{*}{GLM-4.6V-Flash} & $p_{\text{sup}}$ & 0.823 & $\Delta p_{\text{sup}}$ & \cellcolor{green!5}+0.009 & \cellcolor{red!5}-0.004 & \cellcolor{red!5}-0.005 & \cellcolor{red!6}-0.017 & \cellcolor{green!22}+0.056\textsuperscript{***} & \cellcolor{green!5}+0.007 & \cellcolor{green!8}+0.022\textsuperscript{***} \\
  & $p_{\text{ref}}$ & 0.283 & $\Delta p_{\text{ref}}$ & \cellcolor{green!5}+0.013\textsuperscript{*} & \cellcolor{green!5}+0.013 & \cellcolor{green!5}+0.015 & \cellcolor{green!5}+0.006 & \cellcolor{green!32}+0.080\textsuperscript{***} & \cellcolor{green!5}+0.011 & \cellcolor{green!10}+0.026\textsuperscript{***} \\
\addlinespace[2pt]
 \multirow{2}{*}{InternVL3-2B} & $p_{\text{sup}}$ & 0.840 & $\Delta p_{\text{sup}}$ & \cellcolor{red!5}-0.001 & \cellcolor{green!5}+0.002 & \cellcolor{green!5}+0.008 & \cellcolor{red!5}-0.004 & \cellcolor{green!20}+0.051\textsuperscript{***} & \cellcolor{red!5}-0.002 & \cellcolor{green!7}+0.018\textsuperscript{***} \\
  & $p_{\text{ref}}$ & 0.810 & $\Delta p_{\text{ref}}$ & \cellcolor{red!5}-0.013\textsuperscript{*} & \cellcolor{red!5}-0.009 & \cellcolor{red!5}-0.010 & \cellcolor{red!6}-0.017\textsuperscript{*} & \cellcolor{green!16}+0.040\textsuperscript{***} & \cellcolor{red!5}-0.003 & \cellcolor{green!5}+0.014\textsuperscript{*} \\
\addlinespace[2pt]
 \multirow{2}{*}{InternVL3-8B} & $p_{\text{sup}}$ & 0.877 & $\Delta p_{\text{sup}}$ & \cellcolor{green!5}+0.003 & \cellcolor{green!5}+0.007 & \cellcolor{green!5}+0.005 & \cellcolor{red!5}-0.004 & \cellcolor{green!14}+0.037\textsuperscript{***} & \cellcolor{green!5}+0.005 & \cellcolor{green!9}+0.025\textsuperscript{***} \\
  & $p_{\text{ref}}$ & 0.693 & $\Delta p_{\text{ref}}$ & \cellcolor{green!5}+0.006 & +0.000 & \cellcolor{green!5}+0.014 & \cellcolor{green!5}+0.010 & \cellcolor{green!28}+0.071\textsuperscript{***} & \cellcolor{green!5}+0.003 & \cellcolor{green!14}+0.037\textsuperscript{***} \\
\addlinespace[2pt]
 \multirow{2}{*}{InternVL3-38B} & $p_{\text{sup}}$ & 0.816 & $\Delta p_{\text{sup}}$ & \cellcolor{green!5}+0.002 & \cellcolor{red!5}-0.004 & \cellcolor{red!5}-0.002 & \cellcolor{red!10}-0.026\textsuperscript{**} & \cellcolor{green!18}+0.046\textsuperscript{***} & \cellcolor{red!6}-0.017\textsuperscript{**} & \cellcolor{green!5}+0.009 \\
  & $p_{\text{ref}}$ & 0.520 & $\Delta p_{\text{ref}}$ & \cellcolor{green!10}+0.026\textsuperscript{*} & \cellcolor{green!5}+0.007 & \cellcolor{green!10}+0.025\textsuperscript{*} & \cellcolor{green!5}+0.005 & \cellcolor{green!35}+0.090\textsuperscript{***} & \cellcolor{red!5}-0.003 & \cellcolor{green!10}+0.027\textsuperscript{***} \\
\addlinespace[2pt]
 \multirow{2}{*}{Kimi-VL-A3B} & $p_{\text{sup}}$ & 0.834 & $\Delta p_{\text{sup}}$ & \cellcolor{green!5}+0.001 & \cellcolor{red!5}-0.006 & \cellcolor{red!5}-0.000 & \cellcolor{red!5}-0.005 & \cellcolor{green!7}+0.018\textsuperscript{***} & \cellcolor{red!5}-0.002 & \cellcolor{green!5}+0.004 \\
  & $p_{\text{ref}}$ & 0.531 & $\Delta p_{\text{ref}}$ & \cellcolor{red!5}-0.001 & \cellcolor{red!5}-0.014 & \cellcolor{red!5}-0.005 & \cellcolor{red!5}-0.004 & \cellcolor{green!9}+0.024\textsuperscript{***} & \cellcolor{green!5}+0.005 & \cellcolor{green!8}+0.022\textsuperscript{***} \\
\addlinespace[2pt]
 \multirow{2}{*}{Qwen3-VL-2B-Instruct} & $p_{\text{sup}}$ & 0.701 & $\Delta p_{\text{sup}}$ & \cellcolor{red!5}-0.005 & \cellcolor{red!5}-0.009 & \cellcolor{red!5}-0.006 & \cellcolor{red!5}-0.005 & \cellcolor{green!11}+0.029\textsuperscript{***} & \cellcolor{red!5}-0.005 & \cellcolor{green!5}+0.008 \\
  & $p_{\text{ref}}$ & 0.484 & $\Delta p_{\text{ref}}$ & \cellcolor{green!5}+0.010 & \cellcolor{red!5}-0.006 & \cellcolor{green!5}+0.003 & \cellcolor{red!5}-0.000 & \cellcolor{green!18}+0.046\textsuperscript{***} & \cellcolor{green!5}+0.009 & \cellcolor{green!10}+0.025\textsuperscript{***} \\
\addlinespace[2pt]
 \multirow{2}{*}{Qwen3-VL-4B-Instruct} & $p_{\text{sup}}$ & 0.684 & $\Delta p_{\text{sup}}$ & \cellcolor{red!5}-0.001 & \cellcolor{green!5}+0.000 & \cellcolor{red!5}-0.013 & \cellcolor{red!10}-0.026\textsuperscript{*} & \cellcolor{green!22}+0.056\textsuperscript{***} & \cellcolor{green!5}+0.002 & \cellcolor{green!6}+0.017\textsuperscript{*} \\
  & $p_{\text{ref}}$ & 0.282 & $\Delta p_{\text{ref}}$ & \cellcolor{green!5}+0.006 & \cellcolor{red!5}-0.006 & \cellcolor{red!5}-0.006 & \cellcolor{red!9}-0.024\textsuperscript{***} & \cellcolor{green!15}+0.040\textsuperscript{***} & \cellcolor{red!5}-0.008 & \cellcolor{green!7}+0.019\textsuperscript{***} \\
\addlinespace[2pt]
 \multirow{2}{*}{Qwen3-VL-8B-Instruct} & $p_{\text{sup}}$ & 0.679 & $\Delta p_{\text{sup}}$ & \cellcolor{green!5}+0.014\textsuperscript{*} & \cellcolor{red!5}-0.003 & \cellcolor{red!5}-0.006 & \cellcolor{red!14}-0.036\textsuperscript{***} & \cellcolor{green!24}+0.062\textsuperscript{***} & \cellcolor{red!5}-0.007 & \cellcolor{green!6}+0.017\textsuperscript{**} \\
  & $p_{\text{ref}}$ & 0.212 & $\Delta p_{\text{ref}}$ & \cellcolor{green!5}+0.014\textsuperscript{***} & \cellcolor{green!5}+0.005 & \cellcolor{green!5}+0.002 & \cellcolor{red!5}-0.013 & \cellcolor{green!17}+0.045\textsuperscript{***} & \cellcolor{green!5}+0.002 & \cellcolor{green!9}+0.024\textsuperscript{***} \\
\addlinespace[2pt]
 \multirow{2}{*}{Qwen3-VL-32B-Instruct} & $p_{\text{sup}}$ & 0.782 & $\Delta p_{\text{sup}}$ & \cellcolor{green!5}+0.001 & \cellcolor{red!7}-0.018\textsuperscript{**} & \cellcolor{red!10}-0.027\textsuperscript{***} & \cellcolor{red!25}-0.065\textsuperscript{***} & \cellcolor{green!19}+0.049\textsuperscript{***} & \cellcolor{red!6}-0.016\textsuperscript{***} & \cellcolor{green!5}+0.011 \\
  & $p_{\text{ref}}$ & 0.198 & $\Delta p_{\text{ref}}$ & \cellcolor{green!6}+0.015\textsuperscript{***} & \cellcolor{red!5}-0.001 & \cellcolor{green!5}+0.002 & \cellcolor{red!5}-0.012 & \cellcolor{green!24}+0.061\textsuperscript{***} & \cellcolor{red!5}-0.014\textsuperscript{***} & \cellcolor{green!7}+0.019\textsuperscript{***} \\
\bottomrule
\end{tabular}
}
\caption{Changes in $p(\text{support})$ on the supported and refuted subsets after claim rewriting. Values under each rewriting condition represent $\Delta p = p' - p$, where $p$ is the base prediction and $p'$ is the prediction after rewriting. The values of $p(\text{support})$ are computed separately for the supported subset (denoted as $p_{\text{sup}}$) and the refuted subset (denoted as $p_{\text{ref}}$). We observe statistically consistent effects for CT and C1 across models. Significance levels: * $q < 0.05$, ** $q < 0.01$, *** $q < 0.001$ (BH-corrected).
}

\label{tab:p_results}
\end{table*}

The goal of natural rewriting is to demonstrate that even in unintentional, realistic settings, stylistic changes introduced by LLM rewriting still occur and affect downstream tasks. We define five rewriting categories, each reflecting a common motivation for using LLMs to polish academic text:

\begin{itemize}
    \item \textbf{Language Polishing (LP)} Rewrite or fix the text to correct grammatical errors (e.g., \textit{``Can you fix the grammar of this text?''}).
    \item \textbf{Fluency Improvement (FI)} Rewrite the text to improve fluency and readability (e.g., \textit{``Make this paragraph flow more naturally.''}).
    \item \textbf{Academic Style Alignment (AS)} Rewrite the text to match the conventions of a target venue (e.g., \textit{``Rewrite this to match the style of an ACL paper.''}). We select five major ML conferences: ACL, AAAI, NeurIPS, ICLR, and ICML.
    \item \textbf{Outcome-Aware (OA)} Rewrite the text to maximize perceived quality by reviewers, in terms of review scores and acceptance likelihood (e.g., \textit{``Polish this so it sounds more convincing to reviewers.''}).
    \item \textbf{Cautious Tone (CT)} Rewrite the text to make claims more measured and conservative, avoiding overstatement or unsupported certainty (e.g., \textit{``Rewrite this in a more cautious and conservative tone.''}).
\end{itemize}

For each rewriting category, we design five prompt variants for each rewriting category, shown in Table~\ref{tab:rewrite_prompt_new} in the Appendix, to reduce dependence on a single prompt formulation. Instead of applying all variants to every data point, which would increase computational cost by a factor of five, we randomly assign one variant to each data point. This assignment is fixed across all experimental settings to ensure comparability. We also append an additional suffix prompt (see Figure~\ref{fig:natural-rewriting-prompt}) to encourage the model to preserve the underlying meaning of the claim during rewriting.

\paragraph{Human Validation}
\label{sec:human_validation}
To validate the quality of the generated claims, we conducted an annotation
study in which four NLP researchers judged whether rewrites were semantically
consistent with the original, covering 282 claims across conditions OA, CT,
C1, and D1 (Appendix~\ref{sec:human-annotation}). Annotators judged 81.8\% of
rewrites as consistent (from 74.5\% for OA to 90.0\% for D1), though
inter-annotator agreement was low ($\kappa = 0.25$). This low $\kappa$ is
largely a consequence of label skew, and Gwet's AC1, which is robust to skew,
indicates substantial agreement ($\text{AC1} = 0.71$). Together, these suggest
that while annotators agree semantic changes are uncommon, they disagree on
\emph{which} claims cross the boundary.

\subsubsection{Controlled Rewriting}
\label{sec:controlled_rewriting}

While natural rewriting captures realistic use cases, a key observation is that the primary stylistic trait of LLM-rewritten text is the frequent use of characteristic vocabulary. The goal of controlled rewriting is to isolate this effect by selectively injecting LLM-associated words into original claims.

\paragraph{Word sources.}
We draw from two established LLM-vocabulary studies: \citet{doi:10.1126/sciadv.adt3813} and \citet{Matsui2025-nj}. We categorize words into three groups (\textit{decorative}, \textit{conservative}, and \textit{neutral}) based on LLM classification using human-written definitions, reviewed and validated by the author. 

We define two controlled rewriting conditions based on the injected word pool: \textbf{D} for decorative words and \textbf{C} for conservative words. The decorative set serves as a surrogate for rewriting styles such as OA, whereas the conservative set is expected to induce an effect closer to CT. Neutral words are excluded because they are not expected to shift the style signal. This contrast allows us to test whether decorative/conservative vocabulary is a causal factor in downstream performance changes, rather than merely a side effect of the rewriting process. Since neither source contains sufficient hedge words after categorization, we supplement the conservative pool with 31 additional hedging terms identified through LLM-assisted generation. This results in two word pools with 39 conservative words and 114 decorative words.

\paragraph{Word Injection.}
For each claim, the model injects exactly one word from the corresponding pool into the original claim. For each injection, we provide the model with the original claim, the injection instructions, and a randomly sampled subset of candidate words from the selected pool (15 for C, 30 for D). This subsampling encourages diversity across data points while allowing the model to choose a grammatically natural insertion. The prompt requires the model to preserve all original words, avoid deletion, replacement, or reordering, and keep the factual content unchanged (see Figure~\ref{fig:inject-rewriting-prompt}).

\begin{figure}[h]
\centering
\includegraphics[width=\columnwidth]{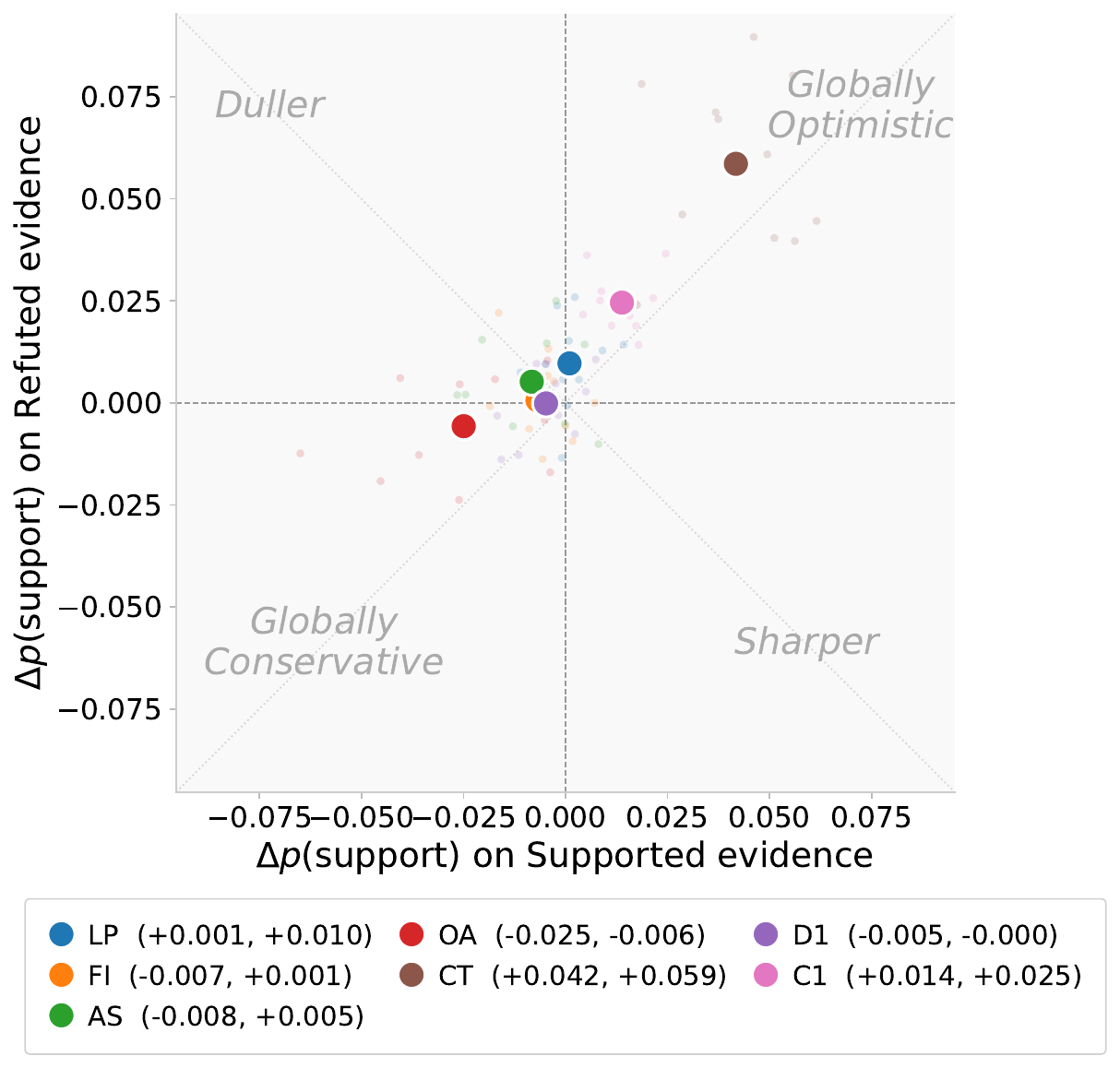}
\caption{Shift decomposition compass. Each point represents one rewriting condition, averaged over all models. The $x$-axis shows the mean change in $p(\text{support})$ on supporting evidence pairs ($\Delta p_{\text{sup}}$) and the $y$-axis on refuting evidence pairs ($\Delta p_{\text{ref}}$). Circles denote different rewriting conditions. Points near the origin indicate robustness to rewriting.}
\label{fig:shift_compass}
\end{figure}

\section{Experimental Setup}
\label{sec:experimental_setup}

\subsection{Datasets}
\label{sec:datasets}

\begin{table}[h]
\centering
\small
\begin{tabular}{lcc}
\toprule
\textbf{Dataset} & \textbf{\#Dev} & \textbf{\#Test} \\
\midrule
\multicolumn{3}{l}{\textit{SciClaimEval}} \\
\quad Full & 747 & 917 \\
\quad Unpaired removed & 704 & 872 \\
\bottomrule
\end{tabular}
\caption{Dataset statistics.}
\label{tab:dataset_stats}
\end{table}

We evaluate on the following multimodal claim verification dataset:
SciClaimEval~\citep{ho2026sciclaimevalcrossmodalclaimverification}. In this dataset, the evidence consists of
table or figure images from scientific papers. We choose to center our analysis on SciClaimEval due to its \emph{paired}
structure: for each claim, an evidence image $e$ is collected and then
modified to construct a refuting counterpart $e'$, resulting in pairs
$(c, e)$ and $(c, e')$ such that $V(c, e) = \textsc{support}$ and
$V(c, e') = \textsc{refute}$. We discard the small number of unpaired
instances arising from slight imbalances in the original dataset. Table~\ref{tab:dataset_stats} summarises the dataset statistics after filtering.

This paired design is critical for our analysis, which modifies claim
text and measures how such modifications affect model predictions. In
standard datasets, the supported and refuted subsets contain different
claims with potentially different stylistic properties, confounding any
modification-based analysis. The paired setting controls for this:
because the same claims appear under both verdicts, the distributional
properties of the two sets of claims are identical, and differences in
outcome can be attributed to the modification itself.

\subsection{Models}
\label{sec:models}

\paragraph{Verification models.}
We select five open-weight VLM families spanning a wide range of
architectures and parameter sizes: Gemma~4~\citep{gemma4},
GLM-4.6V~\citep{hong2025glm}, InternVL3~\citep{zhu2025internvl3exploringadvancedtraining},
Kimi-VL~\citep{kimiteam2025kimivltechnicalreport}, and Qwen3-VL~\citep{bai2025qwen3vltechnicalreport}.
Specifically, we evaluate
Gemma-4-26B-A4B-it and Gemma-4-31B-it,
GLM-4.6V-Flash,
InternVL3-2B, 8B, 38B,
Kimi-VL-A3B-Thinking-2506 (abbreviated as Kimi-VL-A3B),
and Qwen3-VL-2B, 4B, 8B, and 32B-Instruct,
totaling 11 models with parameter counts ranging from 2B to 38B.

\paragraph{Rewriting models.}
Claim rewriting is performed using gpt-oss-120b~\citep{openai2025gptoss120bgptoss20bmodel}, with temperature $T=0.7$ for natural rewriting and $T=0.25$ for controlled injection.

\paragraph{Inference settings.}
Chain-of-thought reasoning is important for solving complex questions such as multimodal claim verification, and methods that extract logprobs without CoT (e.g., forcing the model to output the answer in the first token) yield comparably worse performance. To ensure both strong performance and the ability to extract probability estimates, we employ self-consistency~\citep{wang2023selfconsistencyimproveschainthought} at inference time. For each data point and model, we sample 10 reasoning paths with temperature $T=1.0$ and use the fraction of runs in which the model predicts \textit{support} as $p(\text{support})$.

\paragraph{Computational budgets.}
All experiments were conducted on 8 NVIDIA H200 GPUs over approximately 1–2 days.

\subsection{Metrics.}
\label{sec:metrics}

\paragraph{Paired accuracy and F1.}
We report two levels of performance metrics. F1 score is computed over individual claim-evidence pairs in the standard binary classification sense (support vs.\ refute). Paired accuracy, following the evaluation design of SciClaimEval, applies a stricter criterion: a claim $c$ is paired-correct only if the model produces the correct verdict on \textit{both} the supporting and refuting evidence, i.e., $V(c, e) = \text{support}$ and $V(c, e') = \text{refute}$. We report both metrics before and after rewriting for each condition.

\begin{figure}[h!]
\centering
\includegraphics[width=\columnwidth]{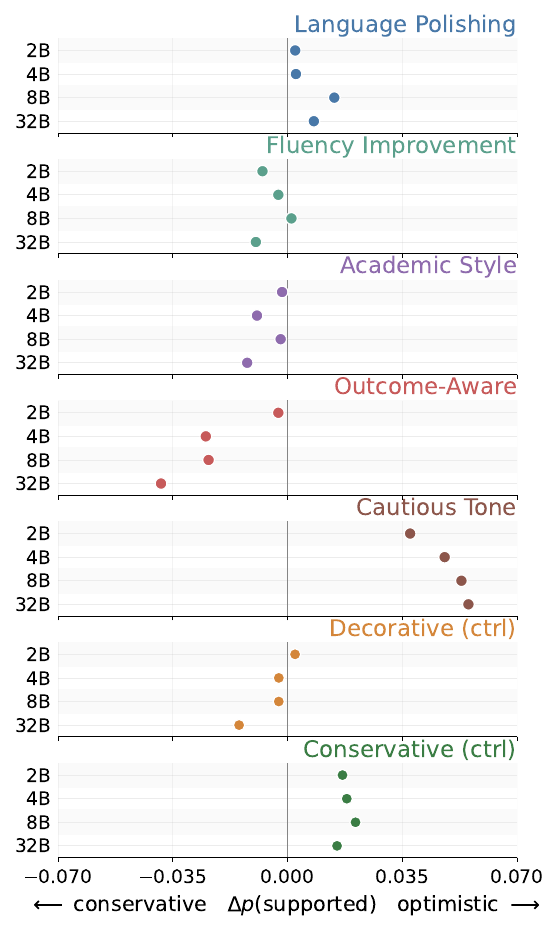}
\caption{Average $\Delta p$ for the \textbf{Qwen3-VL} model family. The x-axis shows the mean change in $p(\text{support})$, i.e., the average predicted probability of the \textit{support} label across the whole dataset. The y-axis represents model size (2B, 4B, 8B, and 32B). }

\label{fig:model_comparison}
\end{figure}

\paragraph{Probability shifts.}

To characterize \textit{how} rewriting affects model behavior, we examine the joint pattern of probability shifts on both pairs. For each claim, we compute:
\begin{align}
    \Delta p_{\text{sup}} &= p(\text{support} \mid e, c') - p(\text{support} \mid e, c) \\
    \Delta p_{\text{ref}} &= p(\text{support} \mid e', c') - p(\text{support} \mid e', c)
\end{align}
The sign pattern of $(\Delta p_{\text{sup}}, \Delta p_{\text{ref}})$ classifies each claim into one of four categories: (1) \textbf{Globally Optimistic} ($\Delta p_{\text{sup}} \uparrow$, $\Delta p_{\text{ref}} \uparrow$), where the model trusts rewritten claims more across the board; (2) \textbf{Globally Conservative} ($\Delta p_{\text{sup}} \downarrow$, $\Delta p_{\text{ref}} \downarrow$), where the model trusts rewritten claims less across the board; (3) \textbf{Sharper} ($\Delta p_{\text{sup}} \uparrow$, $\Delta p_{\text{ref}} \downarrow$), where the model becomes more correct on both pairs; and (4) \textbf{Duller} ($\Delta p_{\text{sup}} \downarrow$, $\Delta p_{\text{ref}} \uparrow$), where the model becomes less correct on both pairs. Figure~\ref{fig:shift_compass} visualises this taxonomy as a two-dimensional compass, placing each rewriting condition in the quadrant. 

\paragraph{Statistical tests.}
We assess the significance of each rewrite condition using a unified paired bootstrap procedure.
For pair accuracy, we compute per-pair differences (condition $-$ base) and bootstrap their mean over 1{,}000 resamples;
for F1, we resample claim indices and recompute $\text{F1}(\text{condition}) - \text{F1}(\text{base})$ on each resample;
for probability shifts $\Delta p$, we bootstrap the mean per-pair shift.
All tests are two-sided: the $p$-value is the fraction of resamples that fall on the minority side of zero, doubled to account for both directions.
Multiple comparisons are corrected by Benjamini--Hochberg FDR~\citep{888cd474-50a6-33fd-a789-415b80e67e78} applied jointly over all (model, condition, metric) cells within each table ($m = 154$ tests), treating the table as the unit of correction. Significance: \sym{*}\,$q < 0.05$, \sym{**}\,$q < 0.01$, \sym{***}\,$q < 0.001$.

\subsection{License.}
The dataset SciClaimEval is released under CC BY 4.0. All models used in this work, including those for rewriting and evaluation, are open-weight and distributed under their respective licenses.

\section{Experimental Results}
\label{sec:results}

\paragraph{RQ1: Does rewriting change the verdict?}
Table~\ref{tab:main_results} reports paired accuracy and F1 across all models and rewriting conditions. We observe few statistically significant changes: the most affected model is gemma-4-31B-it, which shows significant drops under FI, OA, and CT (e.g., 0.610 $\rightarrow$ 0.557, $q < 0.001$ under CT), and InternVL3-8B under CT (0.208 $\rightarrow$ 0.135, $q < 0.001$). These drops are concentrated in a small number of models rather than a consistent trend across all models or strategies.

This finding contrasts with prior work showing that LLM-based rewriting can manipulate review scores~\citep{kaneko2026paraphrasingadversarialattackllmasareviewer, lin-etal-2025-breaking}. A key difference is that those methods employ adversarial strategies designed to influence evaluation outcomes, whereas our rewriting is non-adversarial and does not target any specific metric. Additionally, claim verification is a more constrained setting where the model must ground a single sentence against concrete visual evidence. Whether similar robustness holds under adversarial rewriting optimized for verification remains an open question.

\paragraph{RQ2: Does rewriting change confidence?}
Despite the largely stable accuracy reported in RQ1, Table~\ref{tab:p_results} reveals a much clearer effect at the probability level. The strongest shifts occur under hedging conditions: CT produces significant $\Delta p_{\text{ref}}$ in 11 out of 11 models, and C1 in 11 out of 11, both at $q < 0.05$ or below. In contrast, boosting conditions show a more moderate effect: OA produces significant $\Delta p_{\text{sup}}$ in 6 out of 11 models, but its effect on $\Delta p_{\text{ref}}$ is mixed. One possible explanation is that for refuted claims under boosting conditions, the base $p_{\text{ref}}$ is already low, leaving little room for the probability to decrease further. We observe the same floor-ceiling dynamic in reverse for hedging conditions, where the effect on the supported subset is smaller because base $p_{\text{sup}}$ is already near 1. The remaining conditions (LP, FI, AS) do not produce meaningful probability shifts, with most values small and non-significant. These shifts persist after removing rewrites that insert hedges before numerical or comparative expressions (Appendix~\ref{sec:semantic_change}), and a second rewriter (Gemma-4-31B-it) yields consistent trends (Appendix~\ref{app:rewriter}).

We also do not observe a clear relationship between model size and the magnitude of probability shifts. However, within the Qwen3-VL family (Figure~\ref{fig:model_comparison}), a pattern emerges: larger models show larger drops under OA and larger increases under CT, suggesting that these models become more sensitive to writing style and adjust their predictions accordingly. This pattern is not universal across all model families, so we refrain from drawing a general conclusion.

\paragraph{RQ3: Why do probability shifts not translate to accuracy changes?}
Given the significant probability shifts reported in RQ2, why do accuracy metrics remain largely unchanged? The answer lies in model confidence: most models produce base probabilities clustered near 0 or 1, predicting the same label consistently even across 10 reasoning paths at temperature $T=1.0$. This means that a shift of 0.03--0.05, while statistically significant, is unlikely to cross the 0.5 decision boundary. Table~\ref{tab:flip_rate} confirms this: flip rates are mostly below 15\%. The model's internal belief does shift, but the shift is masked by high confidence. This suggests that on more difficult or subjective cases, where predicted probabilities are closer to 0.5, the effect of stylistic changes on final verdicts may be considerably larger. Representative high-shift and no-shift examples are shown in Appendix~\ref{app:case_studies}.

\section{Conclusion}
In this work, we study how stylistic changes affect model predictions on the multimodal claim verification task. We find that models across different sizes and families are robust to stylistic rewriting in terms of accuracy metrics, but consistent probability shifts exist under certain rewriting conditions. The reason probability shifts occur even when accuracy does not change is that most predictions tend to cluster around decisive points (0 or 1), making the probability changes ineffective when discretized. This is in contrast with previous work on paper review prediction, suggesting that fact-checking tasks are more difficult to sway by simply changing the style of the claims. A natural next question is how this robustness generalizes to other scientific writing tasks beyond reviewing and claim verification. Another interesting direction is to examine how the fact-checking aspect changes when we apply review-score optimization using previously established methods. A better separation of semantic change and stylistic change is also needed to provide more rigorous methodological analysis.

\section{Limitations}

\paragraph{Separating style from semantics.}
Certain rewriting conditions, particularly CT, produce uniquely strong probability shifts, suggesting that the rewriting model sometimes introduces changes that go beyond style.
A common failure mode is the injection of hedging language before numerical values (e.g., ``30\%'' $\rightarrow$ ``roughly 30\%''), which reduces the precision and falsifiability of the claim.
Filtering out such rewrites does not remove the observed shifts (Appendix~\ref{sec:semantic_change}), although a rule-based filter cannot capture all semantic changes.
These cases suggest that fully separating stylistic from semantic modification is inherently difficult: even human annotators often disagree on whether a rewrite preserves meaning (Appendix~\ref{sec:human-annotation}), and even with explicit instructions to preserve meaning, some rewriting scenarios are more disruptive than others.

\paragraph{Single rewriting model and dataset.}
Our main experiments use a single model (gpt-oss-120b) for claim rewriting. A small-scale check with a second rewriter (Gemma-4-31B-it; Appendix~\ref{app:rewriter}) shows consistent trends, but it covers only four conditions, one verifier, and the dev set. Given a fixed computational budget, we prioritize diversity on the \emph{verification} side (11 models across five families), as our primary research question concerns how verifiers respond to rewriting rather than how rewriters differ.
Similarly, we evaluate on a single dataset (SciClaimEval), whose paired structure (identical claims appearing under both supported and refuted evidence) is helpful for our analysis but limits generalizability to other domains and evidence types. To our knowledge, SciClaimEval is the only available dataset that provides this paired design.

\paragraph{Pre-existing LLM polish.} SciClaimEval was constructed from publicly available papers, some of which were published after the widespread adoption of LLMs. It is therefore possible that certain claims in the dataset have already been polished by LLMs in some way, which could dampen the observed effects of our rewriting methods.

\paragraph{Confidence estimation.}
We estimate $p(\text{support})$ only through self-consistency over 10 sampled reasoning paths, which yields probabilities in steps of 0.1. We did not compare it against other confidence extraction methods, such as token-level log-probabilities or verbalized confidence, and the size of the observed shifts may depend on this choice. We leave this comparison to future work.

\section{Ethical Considerations}
\label{sec:ethical_considerations}
This work is an empirical analysis of existing models and datasets, focusing on stylistic changes in scientific writing. All datasets used consist of scientific claims and evidence from published open-access papers and do not contain personally identifying information or offensive content. The dataset (SciClaimEval) is released under CC BY 4.0, and all models used are open-weight and distributed under their respective licenses. The annotation study involved NLP researchers from the research team who evaluated rewritten claims for semantic consistency; no personal or sensitive data was collected. The annotators were co-authors of this paper and were not separately compensated for the annotation task. AI assistants were used for revising author-written text, reviewing ideas, identifying errors in the manuscript, and as a coding assistant for refactoring and implementation. All AI-generated outputs were reviewed and verified by the authors. We acknowledge that while our rewriting strategies are designed to study model robustness, similar techniques could in principle be applied to manipulate automated verification systems. Our code and data are publicly available to support transparency and reproducibility.

\section*{Acknowledgments}
This work was supported by JSPS KAKENHI Grant Number 24K03231. This work was partially funded by the Deutsche Forschungsgemeinschaft (DFG, German Research Foundation) – 
\href{https://gepris.dfg.de/gepris/projekt/554559555}{554559555}. This work was partially supported by the French National Research Agency (ANR) through the FABULEUX project (ANR-26-CE23-6589). F. Boudin was also supported by a visiting professor position at the National Institute of Informatics (NII), Japan. We used ABCI 3.0~\citep{takano2024abci30evolutionleading} provided by AIST and AIST Solutions with support from “ABCI 3.0 Development Acceleration Use”.

\bibliography{anthology,latex/custom}

\appendix

\section{Robustness to the Choice of Rewriter}
\label{app:rewriter}
To test whether our results generalize to other rewriters, we ran an
additional experiment on the OA, CT, C1, and D1 conditions using a different
rewriter model, Gemma-4-31B-it. For this small-scale experiment, we used
Qwen3-VL-8B as the verification model; choosing a model from a different
family than the rewriter avoids self-preference bias. The experiment was
conducted on the dev set only.

\begin{table}[h]
  \centering
  \small
  \setlength{\tabcolsep}{5pt}
  \newcommand{\ci}[2]{{\scriptsize$[#1,\,#2]$}}
  \begin{tabular}{@{}llcc@{}}
    \toprule
    Condition & Rewriter & $\Delta p_{\text{sup}}$ & $\Delta p_{\text{ref}}$ \\
    \midrule
    CT & gpt-oss & $+0.054$ & $+0.030$ \\
       &         & \ci{0.030}{0.078} & \ci{0.012}{0.050} \\[2pt]
       & Gemma   & $+0.056$ & $+0.037$ \\
       &         & \ci{0.032}{0.080} & \ci{0.020}{0.058} \\
    \midrule
    OA & gpt-oss & $-0.035$ & $-0.021$ \\
       &         & \ci{-0.058}{-0.012} & \ci{-0.039}{-0.002} \\[2pt]
       & Gemma   & $-0.027$ & $-0.003$ \\
       &         & \ci{-0.051}{-0.005} & \ci{-0.022}{0.016} \\
    \midrule
    C1 & gpt-oss & $+0.016$ & $+0.018$ \\
       &         & \ci{-0.002}{0.035} & \ci{0.003}{0.033} \\[2pt]
       & Gemma   & $+0.017$ & $+0.023$ \\
       &         & \ci{-0.000}{0.035} & \ci{0.009}{0.038} \\
    \midrule
    D1 & gpt-oss & $-0.003$ & $-0.009$ \\
       &         & \ci{-0.019}{0.013} & \ci{-0.023}{0.005} \\[2pt]
       & Gemma   & $-0.007$ & $-0.004$ \\
       &         & \ci{-0.026}{0.012} & \ci{-0.020}{0.014} \\
    \bottomrule
  \end{tabular}
  \caption{$\Delta p_{\text{sup}}$ and $\Delta p_{\text{ref}}$ for rewrites
  generated by gpt-oss and Gemma-4-31B-it on the dev set, verified by
  Qwen3-VL-8B. Brackets show 95\% bootstrap CIs (1{,}000 resamples). For Gemma, C1/D1 rewrites that do not insert exactly one pool word are excluded.}
  \label{tab:rewriter_robustness}
\end{table}

As shown in Table~\ref{tab:rewriter_robustness}, the direction and magnitude
of the confidence drift in CT, C1, and D1 are broadly similar to those
observed on the gpt-oss rewrites. On OA, the direction holds but the
magnitude is smaller. The difference is clearest on refuted claims, where
the shift essentially vanishes ($-0.021 \rightarrow -0.003$). Overall, this
experiment is consistent with our main finding that hedging-style conditions
produce significant shifts, while boosting conditions show a weaker effect.

\section{Semantic Change of Claims after Rewriting}
\label{sec:semantic_change}
As noted in the Limitations section, the model sometimes produces rewrites
that alter the semantic meaning of the claim. To verify that such rewrites
do not affect our conclusions, we conducted the following analysis. A rewrite
is flagged if it introduces a hedge word (Table~\ref{tab:word-pools})
within the two tokens immediately preceding one of the following elements:
a number, \%, a $p$-value, a unit, or a comparison/ranking word. This
criterion flags 26.3\% of CT rewrites and 22.9\% of C1 rewrites, and fewer
than 10\% of rewrites in all other conditions.

\begin{table}[t]
  \centering
  \small
  \setlength{\tabcolsep}{3.5pt}
  \newcommand{\ci}[2]{{\scriptsize$[#1,\,#2]$}}
  \begin{tabular}{@{}lccc@{}}
    \toprule
    & Full & Unflagged & Flagged \\
    \midrule
    CT $\Delta p_{\text{sup}}$ & $+0.062$ & $+0.055$ & $+0.080$ \\
                               & \ci{0.044}{0.078} & \ci{0.037}{0.074} & \ci{0.051}{0.109} \\[2pt]
    CT $\Delta p_{\text{ref}}$ & $+0.045$ & $+0.044$ & $+0.046$ \\
                               & \ci{0.030}{0.059} & \ci{0.027}{0.061} & \ci{0.020}{0.075} \\
    \midrule
    C1 $\Delta p_{\text{sup}}$ & $+0.017$ & $+0.016$ & $+0.020$ \\
                               & \ci{0.006}{0.028} & \ci{0.003}{0.029} & \ci{-0.002}{0.040} \\[2pt]
    C1 $\Delta p_{\text{ref}}$ & $+0.024$ & $+0.020$ & $+0.040$ \\
                               & \ci{0.014}{0.034} & \ci{0.008}{0.031} & \ci{0.016}{0.063} \\
    \bottomrule
  \end{tabular}
  \caption{$\Delta p_{\text{sup}}$ and $\Delta p_{\text{ref}}$ on the full,
  unflagged, and flagged subsets (Qwen3-VL-8B). Brackets show 95\%
  bootstrap CIs (1{,}000 resamples).}
  \label{tab:flagged_analysis}
\end{table}

As shown in Table~\ref{tab:flagged_analysis}, $\Delta p_{\text{sup}}$ and
$\Delta p_{\text{ref}}$ on the unflagged set are similar to or slightly lower
than those on the full set, while the flagged set tends to show larger deltas
in most conditions. Because the unflagged set preserves the trend observed on
the full set, our conclusions hold even after removing potentially
semantics-altering rewrites.
\section{Prompts}
\label{sec:appendix_prompts}
We provide the full prompt templates used for natural rewriting (Figure~\ref{fig:natural-rewriting-prompt}), controlled rewriting (Figure~\ref{fig:inject-rewriting-prompt}), and the five prompt variants per rewriting category (Table~\ref{tab:rewrite_prompt_new}).

\begin{figure}[h]
\centering
\begin{tcolorbox}[
  promptbox,
  title={Prompt Used for Natural Rewriting}
]
\textbf{[Rewriting Prompt]} During rewriting, you must not alter the factual content of the text. All changes should be purely stylistic and decorative.
\end{tcolorbox}

\caption{Prompt template used for natural rewriting. The \textbf{[Rewriting Prompt]} placeholder is instantiated with one of five rewriting categories: Language Polishing, Fluency Improvement, Academic Style Alignment, Outcome-Aware, or Cautious Tone. See Table~\ref{tab:rewrite_prompt_new} for details.}
\label{fig:natural-rewriting-prompt}
\end{figure}

\section{Flip Rates Across All Models}
\label{sec:appendix_results}
Table~\ref{tab:flip_rate} reports flip rates across all models and rewriting conditions, complementing the probability shift analysis in Section~\ref{sec:results}.

\begin{figure}[t]
\centering
\begin{tcolorbox}[
  promptbox,
  title={Prompt Used for Controlled Rewriting}
]
You are a linguistic editor. Your task is to inject exactly 1 word from the provided word list into the given claim.

\medskip
\noindent\textbf{Rules:}
\begin{itemize}
    \item You MUST insert exactly 1 word from the word list below. This is mandatory.
    \item Do not delete, replace, or reorder any existing words. Every original word must remain.
    \item The injection should be stylistic only---it must not change the factual content.
    \item The injected word must be grammatically natural in its position.
\end{itemize}

\noindent\textbf{Word list:}\\
\texttt{\{word\_list\}}

\medskip
\noindent\textbf{Original claim:}\\
\texttt{\{claim\}}

\medskip
\noindent Output ONLY the modified claim with exactly one word inserted. No explanation, no preamble.
\end{tcolorbox}

\caption{Prompt template used for controlled rewriting. The \texttt{\{word\_list\}} placeholder is instantiated with a randomly sampled subset of candidate words from either the decorative or conservative word pool, and \texttt{\{claim\}} is instantiated with the original claim.}
\label{fig:inject-rewriting-prompt}
\end{figure}

\section{Annotation Guidelines}
\label{sec:appendix_annotation}
Figure~\ref{fig:annotation-guidelines} presents the annotation guidelines provided to annotators for evaluating semantic consistency between original and rewritten claims, as described in Section~\ref{sec:human_validation}.

\section{Word Lists}
\label{sec:appendix_words}
Table~\ref{tab:word-pools} lists the decorative and conservative words used in controlled rewriting, drawn from \citet{doi:10.1126/sciadv.adt3813} and \citet{Matsui2025-nj}.

\section{Data Exclusions}
A small number of claims are excluded from the analysis for two reasons. First, the inject rewriting pipeline (D1, C1) occasionally produces an empty or malformed output; when this affects either member of a claim pair, both members are dropped to preserve the paired evaluation structure (e.g., 29 pairs excluded for D1). Natural rewrite conditions produce no such failures. Second, for a small number of claims, all 10 self-consistency responses from a given model are unparseable; these claims are excluded from that model's evaluation. Crucially, the affected claims are consistent across the baseline and all rewrite conditions for a given model, so no differential bias is introduced into the comparison. In both cases the number of excluded items is small relative to the full evaluation set of 788 pairs.

\section{Human Annotation}
\label{sec:human-annotation}

\begin{table}[h]
  \centering
  \small
  \setlength{\tabcolsep}{5pt}
  \begin{tabular}{@{}lrrrrrr@{}}
    \toprule
    Cond. & Judg. & Pres.\ (\%) & $n_2$ & $P_o$ & $\kappa$ & AC1 \\
    \midrule
    CT  &  96 & 78.1 &  27 & 0.667 & 0.088 & 0.475 \\
    C1  &  97 & 84.5 &  26 & 0.846 & 0.246 & 0.807 \\
    OA  &  98 & 74.5 &  27 & 0.741 & 0.291 & 0.591 \\
    D1  & 100 & 90.0 &  29 & 0.897 & 0.343 & 0.877 \\
    \midrule
    All & 391 & 81.8 & 109 & 0.789 & 0.251 & 0.706 \\
    \bottomrule
  \end{tabular}
  \caption{Human annotation results. Judg.: number of judgments. Pres.:
  percentage of judgments finding the meaning preserved. $n_2$: claims with
  two judgments; $P_o$ (observed agreement), Fleiss' $\kappa$, and Gwet's
  AC1 are computed on these only. CT and C1 are sampled from the unflagged
  subset.}
  \label{tab:human-annotation}
\end{table}

To verify that rewrites preserve the meaning of the original claims, four
annotators, all NLP researchers, answered a single binary question: does the
rewrite keep the same meaning as the original? Annotators saw both claims
side by side with the changed words highlighted, and the rewrite condition
was hidden. The annotators followed the annotation guidelines
(Figure~\ref{fig:annotation-guidelines}). We sampled 80 claims each from OA,
CT, C1, and D1. For CT and C1, we sampled only from the unflagged subset
(Section~\ref{sec:semantic_change}), since this is the subset our analysis
relies on. OA and D1 were sampled at random, since the filter targets
hedging and these are boosting conditions. Each claim was assigned to two annotators; as not all assignments were completed, the returned data contain 391 judgments over 282 claims (109 with two judgments and 173 with one).

Overall, annotators judged the meaning preserved in 81.8\% of judgments,
ranging from 74.5\% (OA) to 90.0\% (D1), with 78.1\% for CT and 84.5\% for
C1 (Table~\ref{tab:human-annotation}). On the 109 doubly annotated claims,
observed agreement is 78.9\%, Fleiss' $\kappa$ is 0.251, and Gwet's AC1 is
0.706. Since each condition contributes only 26--29 doubly annotated claims, per-condition agreement estimates should be interpreted with caution. We use Fleiss' $\kappa$ rather than Cohen's $\kappa$ because
different annotator pairs judged different claims, whereas Cohen's $\kappa$
assumes a single fixed pair of annotators. The low $\kappa$ reflects label
skew: 83.0\% of these judgments mark the meaning as preserved, so $\kappa$
attributes most of the observed agreement to chance, whereas AC1 is robust
to such skew. The remaining disagreement is largely systematic. The rate at
which individual annotators judged a meaning change ranges from 4.7\% to
32.7\%, suggesting that assigning meaning-preservation labels to scientific
claims is inherently difficult. 
\section{Qualitative Examples}
\label{app:case_studies}

To complement the aggregate results, Table~\ref{tab:case-examples} presents representative examples of rewritten claims together with the predictions of Qwen3-VL-8B before and after rewriting.
We include both high-shift cases, where rewriting substantially changes $p(\text{support})$, and no-shift cases, where the prediction remains stable despite the rewrite.

\begin{table*}[t]
\centering
\small
\begin{tabularx}{\textwidth}{@{}lX@{}}
\toprule
\multicolumn{2}{@{}p{\textwidth}@{}}{\textbf{Case 1: high shift} \quad \texttt{val\_fig\_0285} \quad Condition: CT \quad Gold: Supported \quad $p$: 0.0 $\rightarrow$ 0.9 ($\Delta p = +0.9$) \quad Flipped: yes} \\
Original & The results indicated that there were more bacterial categories in clinical samples than cultures by media. \\
Rewritten & The results \textbf{suggested} that \textbf{a greater number of} bacterial categories \textbf{were observed} in clinical samples than \textbf{were identified through culture} media. \\
Evidence & \hspace*{\fill}\raisebox{-0.9\height}{\includegraphics[width=\linewidth,height=3.2cm,keepaspectratio]{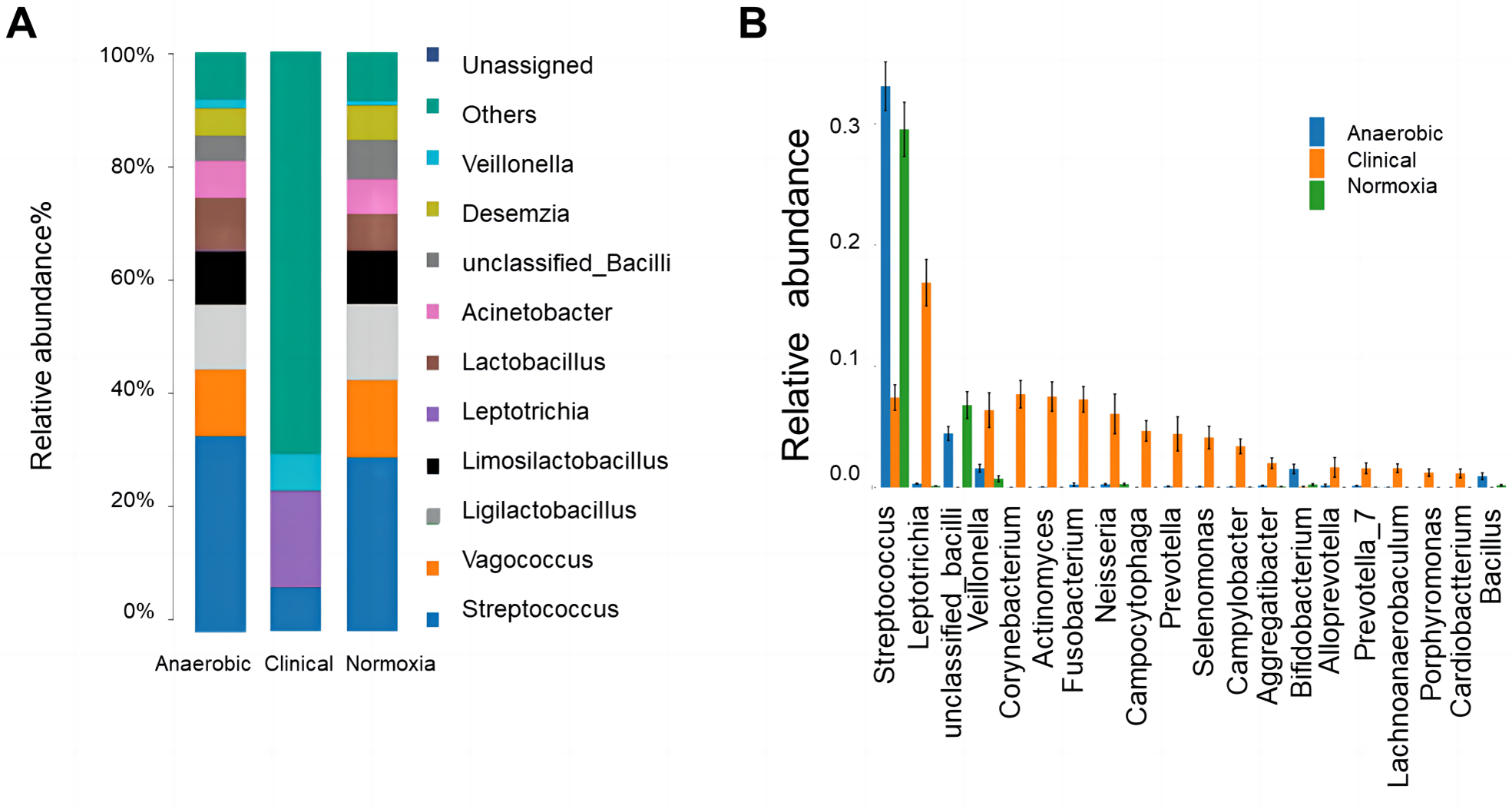}}\hspace*{\fill} \\
\midrule
\multicolumn{2}{@{}p{\textwidth}@{}}{\textbf{Case 2: high shift} \quad \texttt{val\_fig\_0052} \quad Condition: OA \quad Gold: Supported \quad $p$: 0.9 $\rightarrow$ 0.0 ($\Delta p = -0.9$) \quad Flipped: yes} \\
Original & It is also intriguing to observe from Figure 5 that Vanilla LLaMA-2-70B excelled in Writing, but full-parameter fine-tuning led to a decline in these areas, a phenomenon known as the ``Alignment Tax'' (Ouyang et al., 2022 ) . \\
Rewritten & Figure 5 \textbf{compellingly demonstrates} that \textbf{the} vanilla LLaMA-2-70B \textbf{model attains peak performance on writing tasks, yet once the model is subjected to} full-parameter fine-tuning \textbf{its scores markedly decline---a clear manifestation of} the \textbf{well-documented} ``Alignment Tax'' (Ouyang et al., \textbf{2022).} \\
Evidence & \hspace*{\fill}\raisebox{-0.9\height}{\includegraphics[width=\linewidth,height=3.2cm,keepaspectratio]{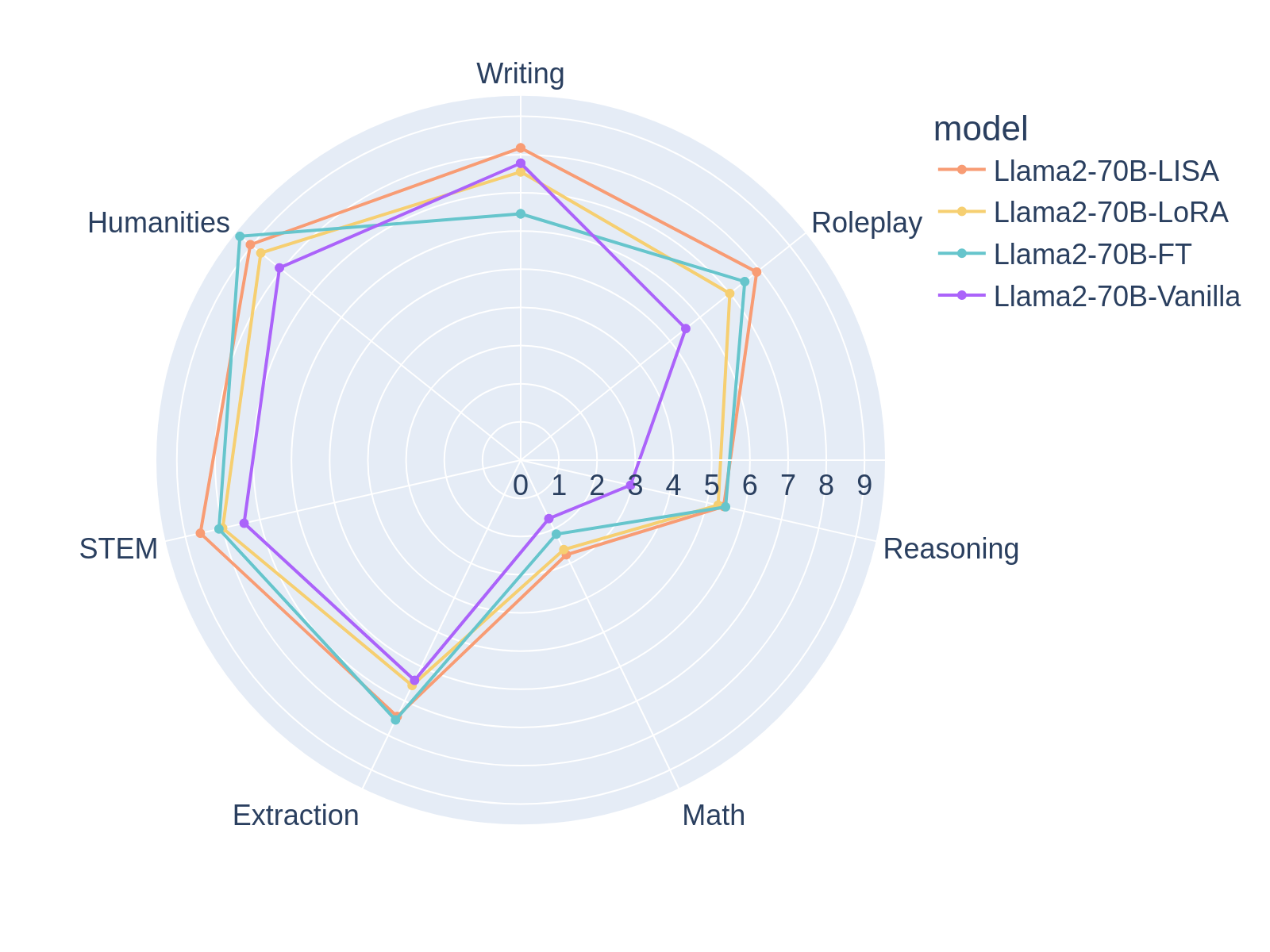}}\hspace*{\fill} \\
\midrule
\multicolumn{2}{@{}p{\textwidth}@{}}{\textbf{Case 3: no shift} \quad \texttt{val\_tab\_0177} \quad Condition: LP \quad Gold: Supported \quad $p$: 0.6 $\rightarrow$ 0.6 ($\Delta p = 0.0$) \quad Flipped: no} \\
Original & Second, pairwise comparisons with the baseline show both CG and finetuning are effective in formality control, where CG has slightly higher win ratio than FT against the baseline. \\
Rewritten & Second, pairwise comparisons with the baseline show \textbf{that} both CG and \textbf{fine-tuning} are effective \textbf{for} formality control, \textbf{with} CG \textbf{achieving a} slightly higher win ratio than FT against the baseline. \\
Evidence & \hspace*{\fill}\raisebox{-0.9\height}{\includegraphics[width=\linewidth,height=3.2cm,keepaspectratio]{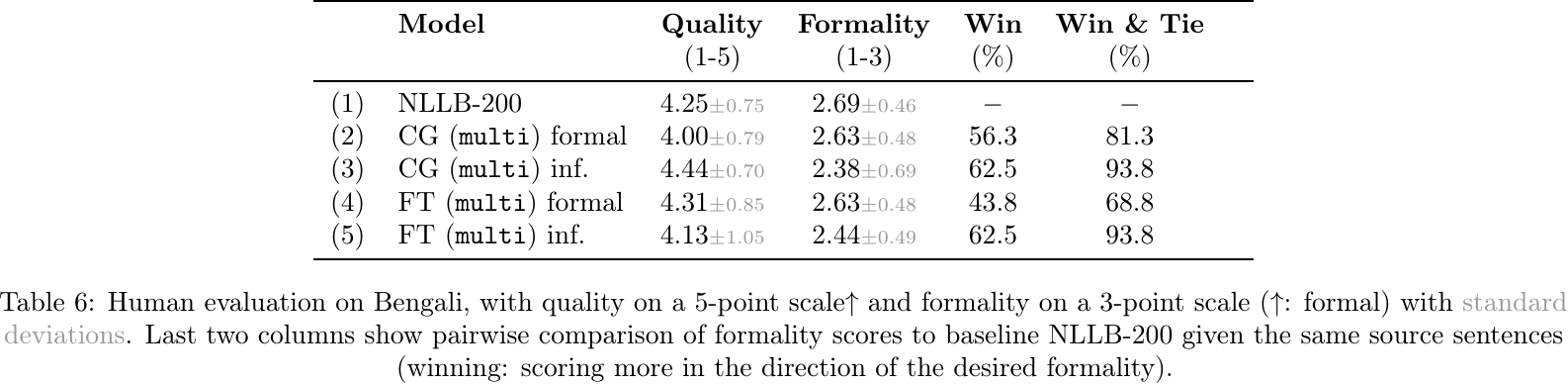}}\hspace*{\fill} \\
\midrule
\multicolumn{2}{@{}p{\textwidth}@{}}{\textbf{Case 4: no shift despite hedging} \quad \texttt{val\_tab\_0029} \quad Condition: C1 \quad Gold: Supported \quad $p$: 0.6 $\rightarrow$ 0.6 ($\Delta p = 0.0$) \quad Flipped: no} \\
Original & Results in Table 1 show that AD-Drop achieves consistent improvement, boosting the average scores of BERT base and RoBERTa base by 0.87 and 0.62, respectively. \\
Rewritten & Results in Table 1 show that AD-Drop achieves \textbf{seemingly} consistent improvement, boosting the average scores of BERT base and RoBERTa base by 0.87 and 0.62, respectively. \\
Evidence & \hspace*{\fill}\raisebox{-0.9\height}{\includegraphics[width=\linewidth,height=3.2cm,keepaspectratio]{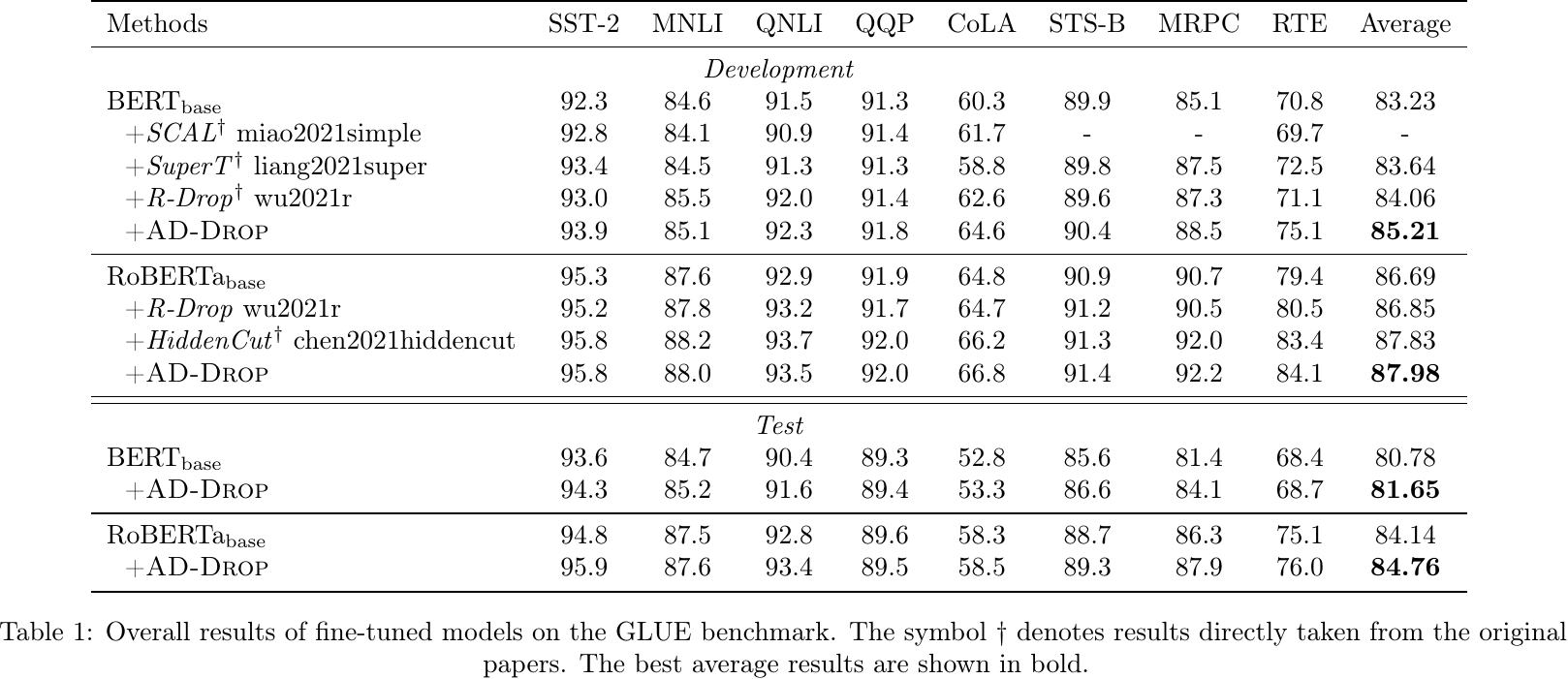}}\hspace*{\fill} \\
\bottomrule
\end{tabularx}
\caption{Example cases for Qwen3-VL-8B-Instruct. $p$ is the probability of \emph{Supported} estimated from 10 sampled answers, before and after rewriting. Words changed by the rewrite are in bold. Evidence images are taken from SciClaimEval \citep{ho2026sciclaimevalcrossmodalclaimverification}.}
\label{tab:case-examples}
\end{table*}

\begin{figure*}[h]
\centering
\begin{tcolorbox}[
  promptbox,
  width=\textwidth,
  title={Annotation Guidelines for Semantic Consistency}
]
\noindent\textbf{I/O:}\\
You will be shown pairs of scientific claims: an original and a rewritten version.

\smallskip
\noindent\textit{Input:} original claim, rewritten claim, evidence image path (for reference)\\
\noindent\textit{Output:} a binary label (TRUE/FALSE), and an optional note field

\medskip
\noindent\textbf{Task:}\\
Does the rewritten claim preserve the same factual meaning as the original?

\begin{itemize}
    \item \textbf{TRUE:} The core factual content is the same. Differences in word choice, tone, emphasis, or formality are fine.
    \item \textbf{FALSE:} The rewrite adds, removes, or changes a factual detail (e.g., a number, a comparison direction, or which method/dataset is referenced).
    \item \textbf{If you can't decide:} Imagine verifying the claim against some evidence (table, figure, etc.). Ask: does there exist any possible evidence that would support the original claim but not the rewritten one, or vice versa? If yes $\rightarrow$ FALSE. If no $\rightarrow$ TRUE.
\end{itemize}

\medskip
\noindent\textbf{Examples:}
\begin{itemize}
    \item ``our method is 20\% better'' $\rightarrow$ ``our method is 25\% better'' $\rightarrow$ \textbf{FALSE} (changed number)
    \item ``our method is 20\% better'' $\rightarrow$ ``our method is 20.0001\% better'' $\rightarrow$ \textbf{FALSE} (even a trivial numerical change could be invalidated by evidence showing exactly 20\%)
    \item ``A is 20\% better than B'' $\rightarrow$ ``A is approximately 20\% better than B'' $\rightarrow$ \textbf{FALSE} (``approximately'' widens the range of valid values)
    \item ``our method is 20\% better'' $\rightarrow$ ``our fantastic method is 20\% better'' $\rightarrow$ \textbf{TRUE} (subjective---no evidence can distinguish ``fantastic'' from ``not fantastic'')
    \item ``consistently outperforms'' $\rightarrow$ ``arguably outperforms'' $\rightarrow$ \textbf{TRUE} (tone shift, but the factual claim is the same)
    \item ``outperforms method A'' $\rightarrow$ ``outperforms method B'' $\rightarrow$ \textbf{FALSE} (different reference)
\end{itemize}
\end{tcolorbox}
\caption{Annotation guidelines provided to annotators for evaluating semantic consistency between original and rewritten claims.}
\label{fig:annotation-guidelines}
\end{figure*}

\begin{table*}[h]
\centering
\begin{tabularx}{\textwidth}{l l X}
\toprule
\textbf{Category} & \textbf{Variant} & \textbf{Prompt} \\
\midrule
\multirow{5}{*}{\shortstack[l]{Language\\Polishing (LP)}}
 & LP-1 & Fix any grammar mistakes in the following text. \\
 & LP-2 & Proofread this text and correct all grammatical errors. \\
 & LP-3 & Can you fix the grammar of this text? \\
 & LP-4 & Please correct any grammatical issues in the text below. \\
 & LP-5 & Review this text for grammar problems and rewrite it with corrections. \\
\midrule
\multirow{5}{*}{\shortstack[l]{Fluency\\Improvement (FI)}}
 & FI-1 & Rewrite this text to make it more fluent and readable. \\
 & FI-2 & Make this paragraph flow more naturally. \\
 & FI-3 & Improve the readability of the following text. \\
 & FI-4 & Smooth out the writing in this text so it reads better. \\
 & FI-5 & Rephrase this text to improve its clarity and flow. \\
\midrule
\multirow{5}{*}{\shortstack[l]{Academic Style\\Alignment (AS)}}
 & AS-1 & Rewrite this to match the writing style expected at ACL. \\
 & AS-2 & Adapt this text to fit the conventions of a AAAI paper. \\
 & AS-3 & I am submitting this paper to NeurIPS. Rewrite it to match the style and quality expected there. \\
 & AS-4 & Revise this text so it reads like a typical ICLR submission. \\
 & AS-5 & Adjust the tone and style of this text to align with papers published at ICML. \\
\midrule
\multirow{5}{*}{\shortstack[l]{Outcome-Aware (OA)}}
 & OA-1 & Rewrite this to be more impressive and compelling to reviewers. \\
 & OA-2 & Polish this so it sounds more convincing to reviewers. \\
 & OA-3 & Rewrite this to enhance the review score and maximize acceptance chances. \\
 & OA-4 & Make this text as persuasive and impactful as possible for peer review. \\
 & OA-5 & Strengthen the writing so that reviewers would strongly favor this paper. \\
 \midrule
\multirow{5}{*}{\shortstack[l]{Cautious\\Tone (CT)}}
 & CT-1 & Rewrite this text to adopt a more cautious and measured tone, softening strong claims and avoiding overgeneralization. \\
 & CT-2 & Rewrite this text to use a more careful and qualified tone, tempering strong assertions and introducing appropriate nuance. \\
 & CT-3 & Revise this passage to emphasize balance and restraint, moderating claims and avoiding overly definitive language. \\
 & CT-4 & Refine this passage to adopt a more reserved academic tone, softening conclusions and avoiding broad or sweeping statements. \\
 & CT-5 & Rewrite this to sound more analytically cautious, reducing emphasis on certainty and acknowledging potential variability in interpretation. \\
\bottomrule
\end{tabularx}
\caption{Prompt strategies for natural rewriting experiments. Each category contains five prompt variants to control for variance introduced by prompt phrasing.}
\label{tab:rewrite_prompt_new}
\end{table*}

\begin{table*}[t]
\centering
\footnotesize
\setlength{\tabcolsep}{4pt}
\resizebox{\textwidth}{!}{%
\begin{tabular}{l l ccccccc}
\toprule
 & & \multicolumn{5}{c}{\textbf{Natural Rewriting}} & \multicolumn{2}{c}{\textbf{Controlled Rewriting}} \\
\cmidrule(lr){3-7} \cmidrule(lr){8-9}
\textbf{Model} & & LP & FI & AS & OA & CT & D1 & C1 \\
\midrule
 \multirow{2}{*}{gemma-4-26B-A4B-it} & Supported & \cellcolor{red!10}4.9\% & \cellcolor{red!10}5.2\% & \cellcolor{red!16}7.9\% & \cellcolor{red!18}8.9\% & \cellcolor{red!16}7.9\% & \cellcolor{red!9}4.5\% & \cellcolor{red!10}5.1\% \\
  & Refuted & \cellcolor{red!13}6.7\% & \cellcolor{red!14}7.0\% & \cellcolor{red!17}8.5\% & \cellcolor{red!18}8.9\% & \cellcolor{red!23}11.5\% & \cellcolor{red!11}5.7\% & \cellcolor{red!11}5.5\% \\
\addlinespace[2pt]
 \multirow{2}{*}{gemma-4-31B-it} & Supported & \cellcolor{red!6}2.9\% & \cellcolor{red!6}3.2\% & \cellcolor{red!8}3.9\% & \cellcolor{red!12}6.1\% & \cellcolor{red!8}4.2\% & \cellcolor{red!6}3.0\% & \cellcolor{red!5}2.4\% \\
  & Refuted & \cellcolor{red!11}5.7\% & \cellcolor{red!13}6.5\% & \cellcolor{red!16}7.9\% & \cellcolor{red!19}9.6\% & \cellcolor{red!24}11.8\% & \cellcolor{red!10}5.1\% & \cellcolor{red!14}7.2\% \\
\addlinespace[2pt]
 \multirow{2}{*}{GLM-4.6V-Flash} & Supported & \cellcolor{red!13}6.7\% & \cellcolor{red!15}7.5\% & \cellcolor{red!17}8.5\% & \cellcolor{red!22}11.0\% & \cellcolor{red!22}10.8\% & \cellcolor{red!10}4.9\% & \cellcolor{red!13}6.5\% \\
  & Refuted & \cellcolor{red!16}7.9\% & \cellcolor{red!19}9.5\% & \cellcolor{red!23}11.4\% & \cellcolor{red!22}10.8\% & \cellcolor{red!31}15.5\% & \cellcolor{red!11}5.5\% & \cellcolor{red!13}6.4\% \\
\addlinespace[2pt]
 \multirow{2}{*}{InternVL3-2B} & Supported & \cellcolor{red!11}5.5\% & \cellcolor{red!11}5.7\% & \cellcolor{red!11}5.6\% & \cellcolor{red!12}6.0\% & \cellcolor{red!9}4.7\% & \cellcolor{red!10}5.1\% & \cellcolor{red!9}4.7\% \\
  & Refuted & \cellcolor{red!15}7.4\% & \cellcolor{red!16}8.0\% & \cellcolor{red!13}6.5\% & \cellcolor{red!17}8.6\% & \cellcolor{red!15}7.7\% & \cellcolor{red!14}7.0\% & \cellcolor{red!12}5.9\% \\
\addlinespace[2pt]
 \multirow{2}{*}{InternVL3-8B} & Supported & \cellcolor{red!10}5.2\% & \cellcolor{red!10}4.9\% & \cellcolor{red!11}5.3\% & \cellcolor{red!10}5.2\% & \cellcolor{red!12}5.8\% & \cellcolor{red!9}4.3\% & \cellcolor{red!6}3.2\% \\
  & Refuted & \cellcolor{red!24}11.9\% & \cellcolor{red!22}10.8\% & \cellcolor{red!23}11.3\% & \cellcolor{red!27}13.6\% & \cellcolor{red!29}14.7\% & \cellcolor{red!22}11.2\% & \cellcolor{red!23}11.5\% \\
\addlinespace[2pt]
 \multirow{2}{*}{InternVL3-38B} & Supported & \cellcolor{red!16}7.8\% & \cellcolor{red!14}7.1\% & \cellcolor{red!16}7.8\% & \cellcolor{red!17}8.7\% & \cellcolor{red!13}6.7\% & \cellcolor{red!13}6.7\% & \cellcolor{red!12}6.0\% \\
  & Refuted & \cellcolor{red!23}11.5\% & \cellcolor{red!22}11.2\% & \cellcolor{red!24}12.2\% & \cellcolor{red!29}14.7\% & \cellcolor{red!30}15.1\% & \cellcolor{red!20}10.0\% & \cellcolor{red!22}11.1\% \\
\addlinespace[2pt]
 \multirow{2}{*}{Kimi-VL-A3B} & Supported & \cellcolor{red!12}6.1\% & \cellcolor{red!20}9.8\% & \cellcolor{red!18}8.9\% & \cellcolor{red!21}10.3\% & \cellcolor{red!16}8.0\% & \cellcolor{red!14}7.0\% & \cellcolor{red!13}6.7\% \\
  & Refuted & \cellcolor{red!23}11.3\% & \cellcolor{red!30}14.8\% & \cellcolor{red!27}13.5\% & \cellcolor{red!29}14.3\% & \cellcolor{red!32}16.1\% & \cellcolor{red!19}9.4\% & \cellcolor{red!23}11.6\% \\
\addlinespace[2pt]
 \multirow{2}{*}{Qwen3-VL-2B-Instruct} & Supported & \cellcolor{red!22}10.9\% & \cellcolor{red!22}10.9\% & \cellcolor{red!23}11.3\% & \cellcolor{red!27}13.7\% & \cellcolor{red!31}15.4\% & \cellcolor{red!26}12.8\% & \cellcolor{red!23}11.4\% \\
  & Refuted & \cellcolor{red!25}12.3\% & \cellcolor{red!34}16.8\% & \cellcolor{red!32}15.9\% & \cellcolor{red!33}16.7\% & \cellcolor{red!31}15.3\% & \cellcolor{red!23}11.7\% & \cellcolor{red!25}12.7\% \\
\addlinespace[2pt]
 \multirow{2}{*}{Qwen3-VL-4B-Instruct} & Supported & \cellcolor{red!23}11.3\% & \cellcolor{red!23}11.6\% & \cellcolor{red!24}12.1\% & \cellcolor{red!30}14.8\% & \cellcolor{red!30}15.1\% & \cellcolor{red!16}7.9\% & \cellcolor{red!19}9.7\% \\
  & Refuted & \cellcolor{red!22}10.8\% & \cellcolor{red!20}9.9\% & \cellcolor{red!24}12.1\% & \cellcolor{red!25}12.7\% & \cellcolor{red!25}12.7\% & \cellcolor{red!17}8.6\% & \cellcolor{red!18}8.8\% \\
\addlinespace[2pt]
 \multirow{2}{*}{Qwen3-VL-8B-Instruct} & Supported & \cellcolor{red!19}9.3\% & \cellcolor{red!21}10.4\% & \cellcolor{red!24}12.1\% & \cellcolor{red!30}14.8\% & \cellcolor{red!27}13.5\% & \cellcolor{red!16}7.8\% & \cellcolor{red!17}8.7\% \\
  & Refuted & \cellcolor{red!15}7.6\% & \cellcolor{red!17}8.3\% & \cellcolor{red!18}8.8\% & \cellcolor{red!20}10.2\% & \cellcolor{red!24}11.8\% & \cellcolor{red!16}8.2\% & \cellcolor{red!14}7.2\% \\
\addlinespace[2pt]
 \multirow{2}{*}{Qwen3-VL-32B-Instruct} & Supported & \cellcolor{red!12}6.0\% & \cellcolor{red!12}6.2\% & \cellcolor{red!18}8.9\% & \cellcolor{red!21}10.7\% & \cellcolor{red!18}9.0\% & \cellcolor{red!13}6.3\% & \cellcolor{red!14}6.9\% \\
  & Refuted & \cellcolor{red!14}7.0\% & \cellcolor{red!13}6.6\% & \cellcolor{red!16}8.0\% & \cellcolor{red!20}9.8\% & \cellcolor{red!21}10.7\% & \cellcolor{red!12}5.9\% & \cellcolor{red!14}7.2\% \\
\bottomrule
\end{tabular}
}
\caption{Flip rate: percentage of claims whose predicted label (threshold $0.5$ on $p(\text{support})$) changes after rewriting, in either direction. Rows group claims by gold label: \textit{Supported} (claim paired with supporting evidence) and \textit{Refuted} (claim paired with refuting evidence). Colour intensity scales with flip rate.
}

\label{tab:flip_rate}
\end{table*}

\begin{table*}[h]
\centering
\begin{tabularx}{\textwidth}{l X}
\toprule
\textbf{Category} & \textbf{Words} \\
\midrule
\multirow{7}{*}{\shortstack[l]{Conservative (C)}}
 &    acknowledges, allegedly, apparently, approximately, arguably,
   basically, broadly, comparatively, conceivably, essentially,
   generally, hinting, inadequately, largely, mainly, marginally,
   moderately, mostly, often, ostensibly, partially, perhaps,
   plausibly, possibly, potential, presumably, primarily, purportedly,
   relatively, roughly, seemingly, slightly, somewhat, supposedly,
   tentatively, typically, underexplored, unexplored, usually \\
\midrule
\multirow{18}{*}{\shortstack[l]{Decorative (D)}}
 &    accentuates, adept, advocates, affirming, aptly, attains,
   augmenting, avenue, boast, bolster, burgeoning, capitalizing,
   catalyze, combating, commendable, compelling, compellingly,
   comprehensive, consolidates, crafted, critical, crucial,
   culminating, deeper, dependability, dependable, detrimentally,
   disrupts, distinctive, effectively, effortlessly, elevate,
   emphasises, emphasising, emphasize, empowers, enduring, enhance,
   essential, excel, excellently, exceptional, exceptionally,
   exhaustive, expansive, formidable, fortify, foster, foundational,
   fresh, fundamental, groundbreaking, illuminate, imperative,
   impressive, impressively, ingenious, innovative, intriguing,
   invaluable, leverage, lucidly, meticulous, meticulously,
   multifaceted, notable, notably, noteworthy, optimizing,
   orchestrating, outperform, paving, pinpoint, pioneering, pivotal,
   poised, potent, pressing, promise, pronounced, propelling,
   remarkable, renowned, revolutionize, seamless, seamlessly, showcase,
   significant, solidify, strategically, streamline, substantiated,
   surged, surmount, surpass, swift, swiftly, thorough, transcend,
   transform, transformative, uncharted, uncovering, underscore,
   unlocking, unparalleled, unraveling, unveil, uphold, valuable,
   versatile, versatility, warranting, well-rounded \\
\bottomrule
\end{tabularx}
\caption{Word pools used in inject rewriting. These words are frequently
included by LLMs during writing and serve as recognizable markers of
LLM-generated text. Decorative words are sourced from
\citet{doi:10.1126/sciadv.adt3813} and \citet{Matsui2025-nj};
conservative words are drawn from the same sources and supplemented with
LLM-assisted hedging terms. Category C (Conservative) contains 39 words;
Category D (Decorative) contains 114 words.}
\label{tab:word-pools}
\end{table*}

\end{document}